\documentclass[10pt,twocolumn,letterpaper]{article}

\usepackage[pagenumbers]{cvpr} % To force page numbers, e.g. for an arXiv version

\usepackage[dvipsnames]{xcolor}

\newcommand{\pipelinename}{\textit{Gen4Gen}\xspace}
\newcommand{\datasetname}{\textit{MyCanvas}\xspace}

\usepackage[pagebackref,breaklinks,colorlinks,citecolor=cvprblue]{hyperref}

\def\paperID{****} % *** Enter the Paper ID here
\def\confName{CVPR}
\def\confYear{2024}

\usepackage{amsmath,amsfonts,bm}

\usepackage{bbm}
\usepackage{bm}

\usepackage{booktabs}       % professional-quality tables
\usepackage{amsfonts}       % blackboard math symbols
\usepackage{nicefrac}       % compact symbols for 1/2, etc.
\usepackage{microtype}      % microtypography
\usepackage{color}
\usepackage{array}
\usepackage{multirow}
\usepackage{dirtytalk}
\usepackage{amsthm}
\usepackage[para,online,flushleft]{threeparttable}
\usepackage{colortbl}
\usepackage[normalem]{ulem}

\definecolor{Crimson}{rgb}{0.86, 0.08, 0.24}
\definecolor{DarkGreen}{rgb}{0.10, 0.55, 0.10}
\definecolor{RoyalBlue}{rgb}{0.20, 0.60, 0.86}
\definecolor{DarkCyan}{rgb}{0.0, 0.54, 0.54}
\definecolor{Gray}{gray}{0.9}
\definecolor{ChromeYellow}{rgb}{1.0, 0.65, 0.0}
\definecolor{Salmon}{rgb}{0.98, 0.50, 0.45}
\definecolor{LightGreen}{rgb}{0.93,0.98,0.96}
\ifx \submission \undefined

\else
\fi

\usepackage[most]{tcolorbox}
\newtcolorbox{prompt}{
    colback=gray!10,
    colframe=gray!40,
    fonttitle=\bfseries,
    coltitle=black,
    colbacktitle=gray!40,
    enhanced,
    drop shadow=black!5!white,
    left=8mm,
    right=8mm,
    top=3mm,
    bottom=3mm,
    boxsep=1mm,
    title=Text Prompt:
    }
\newtcolorbox{fs_template}{
    colback=gray!10,
    colframe=gray!40,
    fonttitle=\bfseries,
    coltitle=black,
    colbacktitle=gray!40,
    enhanced,
    drop shadow=black!5!white,
    left=8mm,
    right=8mm,
    top=3mm,
    bottom=3mm,
    boxsep=1mm,
    title=Few-Shot Template:
    }
\newtcolorbox{fs_query}{
    colback=gray!10,
    colframe=gray!40,
    fonttitle=\bfseries,
    coltitle=black,
    colbacktitle=gray!40,
    enhanced,
    drop shadow=black!5!white,
    left=8mm,
    right=8mm,
    top=3mm,
    bottom=3mm,
    boxsep=1mm,
    title=Few-Shot Queries:
    }
\newtcolorbox{bkg_template}{
    colback=gray!10,
    colframe=gray!40,
    fonttitle=\bfseries,
    coltitle=black,
    colbacktitle=gray!40,
    enhanced,
    drop shadow=black!5!white,
    left=8mm,
    right=8mm,
    top=3mm,
    bottom=3mm,
    boxsep=1mm,
    title=Few-Shot Background Template:
    }
\newtcolorbox{bkg_query}{
    colback=gray!10,
    colframe=gray!40,
    fonttitle=\bfseries,
    coltitle=black,
    colbacktitle=gray!40,
    enhanced,
    drop shadow=black!5!white,
    left=8mm,
    right=8mm,
    top=3mm,
    bottom=3mm,
    boxsep=1mm,
    title=Few-Shot Background Queries:
    }
\newtcolorbox{bkg_prompt}{
    colback=gray!10,
    colframe=gray!40,
    fonttitle=\bfseries,
    coltitle=black,
    colbacktitle=gray!40,
    enhanced,
    drop shadow=black!5!white,
    left=8mm,
    right=8mm,
    top=3mm,
    bottom=3mm,
    boxsep=1mm,
    title=Retrieved Background:
    }
\begin{document}

\definecolor{cvprblue}{rgb}{0.21,0.49,0.74}

%%%%%%%%% TITLE - PLEASE UPDATE
\title{
\raisebox{-0.35ex}{\protect\includegraphics[height=1.5\fontcharht\font`\B]{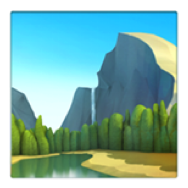}}
\pipelinename: Generative Data Pipeline for Generative Multi-Concept Composition}

%%%%%%%%% AUTHORS - PLEASE UPDATE
\author{Chun-Hsiao Yeh\textsuperscript{1*} \quad 
Ta-Ying Cheng\textsuperscript{2*} \quad 
He-Yen Hsieh\textsuperscript{3*} \quad 
Chuan-En Lin\textsuperscript{4} \quad 
Yi Ma\textsuperscript{1,5} \\
Andrew Markham\textsuperscript{2} \quad 
Niki Trigoni\textsuperscript{2} \quad 
H.T. Kung\textsuperscript{3} \quad 
Yubei Chen\textsuperscript{6}{$^\dagger$} \quad \\
\textsuperscript{1}UC Berkeley \quad
\textsuperscript{2}University of Oxford \quad
\textsuperscript{3}Harvard University \quad
\textsuperscript{4}CMU \quad
\textsuperscript{5}HKU \quad
\textsuperscript{6}UC Davis
}

\twocolumn[{
\renewcommand\twocolumn[1][]{#1}%
\maketitle
\vspace{-10mm}
\begin{center}
    \centering
    \includegraphics[width=0.95\linewidth]{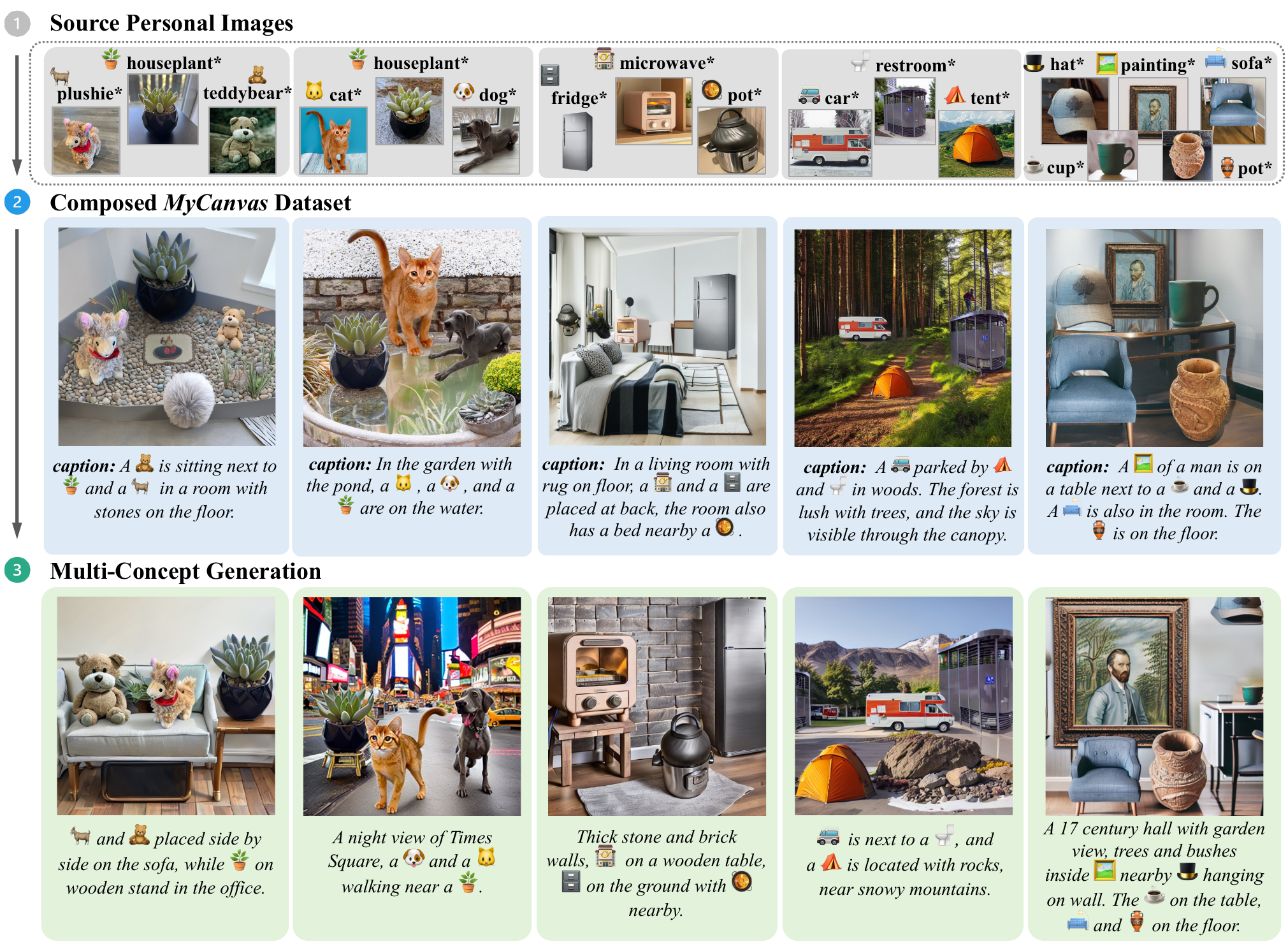}
    \vspace{-9pt}
    \captionof{figure}{Given very few source images representing several concepts (each denoted by a \textbf{concept*}), we introduce a semi-automated dataset creation pipeline, \pipelinename, to compose these \textbf{concept*} into realistic scenes with complex compositions, accompanied by detailed text descriptions, namely, \datasetname. Using this composed \datasetname dataset boosts the performance of previous methods in multi-concept personalization without amending the architecture or training algorithms. Our \datasetname dataset addresses issues in prior works that fail to extend to multiple concept generation (beyond three concepts) or challenging cases (e.g., dog and cat, teddybear and plushie).}
    \label{fig:teaser}
\end{center}
}]

{\let\thefootnote\relax\footnote{$^*$Equal Contribution $^\dagger$Corresponding Author}}

\begin{abstract}

Recent text-to-image diffusion models are able to learn and synthesize images containing novel, personalized concepts (e.g., their own pets or specific items) with just a few examples for training. This paper tackles two interconnected issues within this realm of personalizing text-to-image diffusion models. First, current personalization techniques fail to reliably extend to multiple concepts --- we hypothesize this to be due to the mismatch between complex scenes and simple text descriptions in the pre-training dataset (e.g., LAION). Second, given an image containing multiple personalized concepts, there lacks a holistic metric that evaluates performance on not just the degree of resemblance of personalized concepts, but also whether all concepts are present in the image and whether the image accurately reflects the overall text description. To address these issues, we introduce \pipelinename, a semi-automated dataset creation pipeline utilizing generative models to combine personalized concepts into complex compositions along with text-descriptions. Using this, we create a dataset called \datasetname, that can be used to benchmark the task of multi-concept personalization. In addition, we design a comprehensive metric comprising two scores (CP-CLIP and TI-CLIP) for better quantifying the performance of multi-concept, personalized text-to-image diffusion methods. We provide a simple baseline built on top of Custom Diffusion with empirical prompting strategies for future researchers to evaluate on \datasetname. We show that by improving data quality and prompting strategies, we can significantly increase multi-concept personalized image generation quality, without requiring any modifications to model architecture or training algorithms. We demonstrate that chaining strong foundation models could be a promising direction for generating high-quality datasets targeting a variety of challenging tasks in the computer vision community. The project is available at \url{https://danielchyeh.github.io/Gen4Gen/}.

\end{abstract}    
\section{Introduction}

From photorealistic portraits to paintings of fantasy creatures, the past year has seen a remarkable leap in the capabilities of text-to-image diffusion models~\cite{nichol2021glide,ramesh2022hierarchical,gu2022vector,rombach2022high,saharia2022photorealistic,podell2023sdxl,dalle3}. Some recent efforts have focused on the ``personalization" of these generative models, wherein a pre-trained text-to-image diffusion model is augmented with a minimal set of user-provided concept images (e.g., their pets or recently bought houseplant) to generate new scenes incorporating these personal concepts (e.g., their pets in a night view of Times Square as shown in Figure~\ref{fig:teaser}).
Notable works in this area~\cite{ruiz2023dreambooth,kumari2023multi,Han_2023_ICCV,alaluf2023neural,ma2023subject} are important milestones as they increase users' control over the generation process, bringing out a variety of tailored applications.

However, it can be challenging to perform personalization on multiple concepts simultaneously and control the image generation to accurately follow the given text descriptions. Moreover, ~\cite{kumari2023multi} points out that even in general cases, stable diffusion~\cite{rombach2022high} fails to disentangle and present multiple concepts in the same image when their latent spaces are similar (e.g., dog and cat). This issue is often inherited to the subsequently fine-tuned personalization models. We conjecture that this behavior was the result of a mismatch between the text-image pairs in the pre-training dataset (e.g., LAION~\cite{schuhmann2022laion}). Many images in LAION are often single-object-centric, with the accompanying captions providing a broad overview of the scenes rather than offering detailed descriptions of individual concepts. The lack of correspondence between the text and complex image compositions poses a challenge in generating multiple concepts, especially when concepts are semantically similar.

To validate our hypothesis that better data quality would lead to better multi-concept personalization, we decide to go down a different route contrary to previous model-driven techniques, tackling this problem by constructing a proof-of-concept dataset for personalization with multi-concept-centric images and text descriptions. To do so, we leverage the recent advancements in highly accurate foundation models, and introduce a semi-automated generative data pipeline for the composition of multiple personlized concepts; We hence call our approach, \pipelinename. This dataset creation pipeline leverages the recent advancements in image foreground extractions~\cite{qin2022}, Large Language Models (LLMs)~\cite{lian2023llmgrounded}, image inpainting~\cite{podell2023sdxl}, and Multimodal Large Language Models (MLLMs)~\cite{liu2023visual}, to re-composite sets of user-provided photos into realistic, personalized multi-concept images with densely corresponding text descriptions. In addition, we dive into the realm of prompt engineering to further improve the caption quality during training time for better image-text alignment. We generate and filter over 10k images and create the final benchmarking dataset \datasetname. 

In the process of creating a better benchmark, we also realize the importance of a reasonable evaluation metric that could be applied to all personalization fine-tuning methods, given that most benchmarks~\cite{ruiz2023dreambooth,kumari2023multi,Han_2023_ICCV,huang2023t2icompbench,bakr2023hrs} focus on either the more general case of generalization or evaluate only up to three personalized concepts with heavy amount of the comparisons relying on user surveys. Thus, we draw inspiration from the taxonomy of ~\cite{saharia2022photorealistic,cho2023dall,petsiuk2022human,dinh2022tise,huang2023t2icompbench,bakr2023hrs} and present a Composition-Personalization-CLIP score (CP-CLIP) and a Text-Image alignment CLIP score (TI-CLIP). The two scores act as a simple yet holistic metric that takes into both composition and personalization accuracy and the ability to generalize to various scenarios.

We show that previous methods~\cite{ruiz2023dreambooth,kumari2023multi}, with an enhanced dataset like \datasetname and our prompting strategy, can gain significant improvements in generating realistic multi-concept images with varying backgrounds while sticking to the identity of personalized concepts. The improvements are even more apparent under very complex compositions, challenging guidance (relative positions), and multiple semantically similar concepts (e.g., two dog identities in the same image). The promising results gained from the improvement of dataset quality using our semi-automated data creation approach are a motivation for opportunities in chaining AI foundation models to create large-scale high-quality datasets in the near future.

Overall, our paper contributes three important findings: \textbf{(\textit{i}) Integrating AI foundation models is crucial:} The semi-automated dataset creation pipeline, \pipelinename, introduces the possibilities of using cascaded AI-foundation models for generating high-quality datasets, and holds the promise for benefiting a wide range of tasks. \textbf{(\textit{ii}) Dataset quality matters:} Our proof-of-concept \datasetname dataset is a reflection that simply composing well-aligned image and text description pairs would significantly improve the task of multi-concept personalization. \textbf{(\textit{iii}) A benchmark for multi-concept personalization is required:} Our holistic evaluation benchmark considers personalization accuracy, composition correctness, and text-image alignment in the task of multi-concept personalization. We hope our \datasetname dataset along with the metric, CP-CLIP and TI-CLIP scores can be used as a better measure to address this purpose.

\section{Related Works}

\noindent\textbf{Personalized Text-to-Image Generation.}
Given a pre-trained text-to-image diffusion model and very few user-provided images representing a specific concept, the goal of personalization is to fine-tune the model and find a special identifier mapping to the concept. The identifier is then used to generate new scenes incorporating the specific concept in it. Textual Inversion~\cite{gal2022image} and DreamBooth ~\cite{ruiz2023dreambooth} are the first few to tackle this task. The former learns a token embedding to create the mapping without changing the model, while the latter fine-tunes the entire model while ensuring the generalization capability still remains. Many works soon followed, focusing on both the fidelity and identity preservation of the object~\cite{alaluf2023neural,chen2023suti,chen2023disenbooth,Wei_2023_ICCV,chen2023photoverse,wu2023singleinsert}, and extending into multi-concept personalization~\cite{kumari2023multi,Han_2023_ICCV,liu2023cones,gu2023mix,xiao2023fastcomposer,avrahami2023break}. Specifically, these methods focused on using regularized finetuning to improve the generalization given data scarcity, e.g., Custom Diffusion~\cite{kumari2023multi} fine-tunes only K and V cross-attention layers which potentially leads to less overfitting, SVDiff~\cite{Han_2023_ICCV} optimizes the singular values of weights, Cones2~\cite{liu2023cones} learns a residual embedding to shift general concepts to personalized ones. On the other hand, we aim to target the same problem with a data-centric approach, showing that simply improving the dataset could lead to significant performance leaps in terms of multi-concept personalization.

\noindent\textbf{Text-to-Image Datasets and Benchmarks.}
The main ignition behind the success of diffusion models lies in their vast quantity of data~\cite{rombach2022high,saharia2022photorealistic,podell2023sdxl,dalle3}. The text-encoder and subsequently the diffusion model itself are often built on datasets in the scale of billions~\cite{schuhmann2022laion,radford2021learning,raffel2020exploring}. Inevitably, a lot of data within the dataset are of poor quality, particularly in terms of the alignment between text and image, where a complex scene may only be described by a few words~\cite{schuhmann2022laion}. Concurrent works such as DALLE-3~\cite{dalle3} and RECAP~\cite{segalis2023picture} also observed this similar phenomenon. Our work aims to show that a proof-of-concept personalization dataset comprising multiple concepts in the same scene and well-aligned text descriptions can improve the process of fine-tuning even when the quantity of data is small.
Another open challenge for these generative models is how to evaluate them holistically. Recently, works like DrawBench~\cite{saharia2022photorealistic}, T2I-CompBench~\cite{huang2023t2icompbench}, and HRS~\cite{bakr2023hrs} have provided a more holistic approach in evaluating these text-to-image diffusion models. We draw inspiration from them to propose the first holistic benchmark in evaluating the specific task of multi-concept personalization.

\section{\pipelinename: A Data-Driven Approach to Multi-Concept Personalization}

\begin{figure*}[ht]
    \centering
    \includegraphics[width=\linewidth]{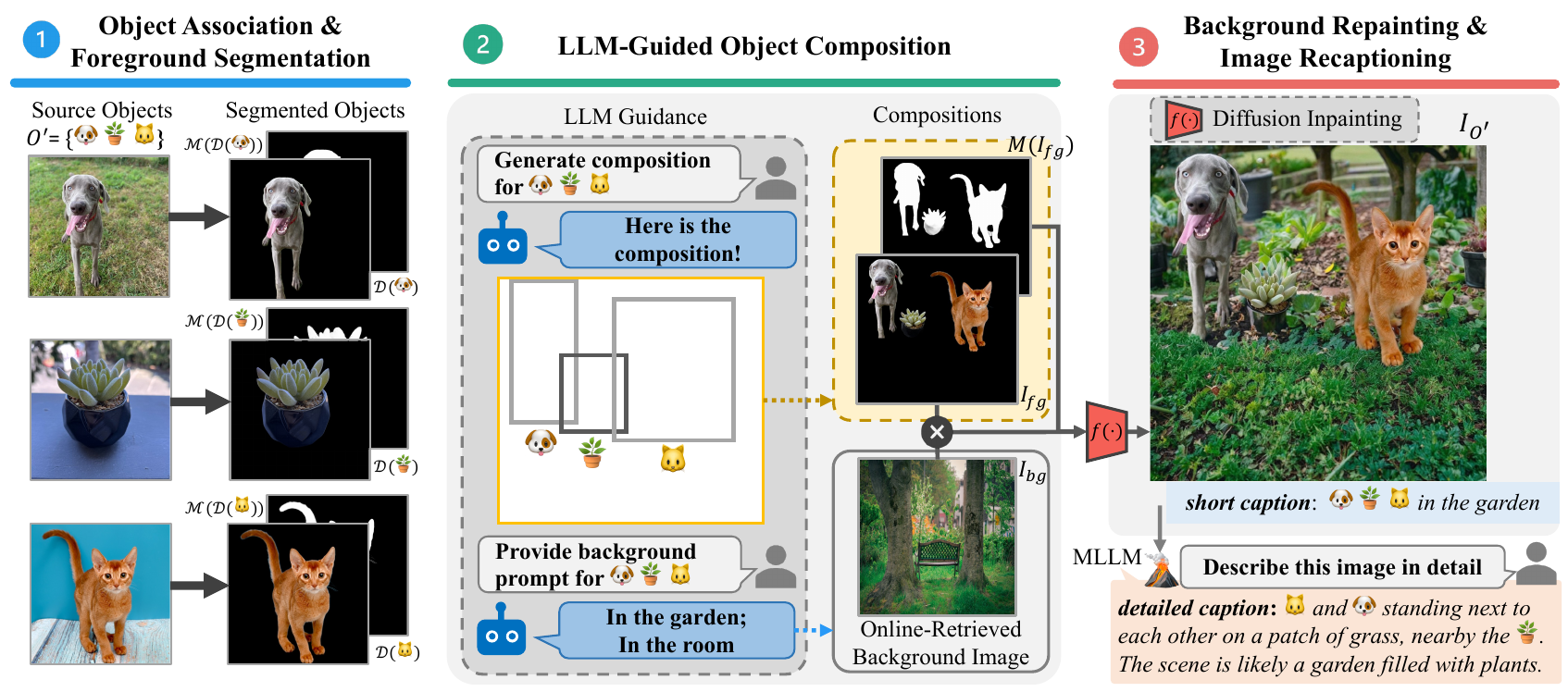}
    \vspace{-10pt}
    \caption{\textbf{Overview of our \pipelinename Pipeline for Creating \datasetname Dataset.} Given source images representing multiple concepts, we leverage recent advancements in image foreground extraction, LLMs, MLLMs, and inpainting to compose realistic, personalized images and paired text descriptions. Our \pipelinename pipeline has three stages. First (1), we apply a category-agnostic saliency object detector to segment the foreground given objects within composition $O'$. Second (2), we inquire the LLM to provide probable bounding box compositions. This forms the composite foreground image $I_{fg}$ and its corresponding mask $\mathcal{M}(I_{fg})$. In addition, we ask the LLM to provide a set of background prompts describing potential scenes for $O'$. Third (3), we use a diffusion inpainting model to repaint $I_{fg}$ by embedding it within a background image $I_{bg}$ sourced from the internet to produce the final image $I_{O'}$. To increase the variety of text descriptions while maintaining the alignment, we then inquire a MLLM (LLaVA) to provide a detailed caption describing $I_{O'}$ to a subset of all combinations.
     }
    \label{fig:data-creation}
    \vspace{-5pt}
\end{figure*}

Given a set of user-provided photos capturing multiple concepts (e.g., dog and houseplant), the goal of multi-concept personalization is to learn the identity of each concept such that we can synthesize new images comprising multiple of them with varying backgrounds and compositions. As shown in previous approaches~\cite{kumari2023multi}, the problem difficulty increases significantly with the number of personalized concepts we aim to inject into an image.

While previous works~\cite{kumari2023multi,Han_2023_ICCV,liu2023cones,Wei_2023_ICCV,gu2023mix,xiao2023fastcomposer} have focused on optimizing training strategies, in this paper, we demonstrate that improving data quality throughout the training enhances the generation quality of multi-concept personalized images.

In this section, we will discuss our main contributions to the advancements of multi-concept personalization.
\subsection{Dataset Design Principles}
\label{sec:laion}
From the most aesthetic subset (\textit{LAION-2B-en improved Aesthetics}) of the LAION dataset~\cite{schuhmann2022laion}, we can clearly see a mismatch between the complexity of an image against the simple descriptions. Since the dataset is largely web-retrieved, discrepancies may arise. For example, there could exist inaccurate and abundant text descriptions for an image, as well as low resolutions over images containing multiple objects (see Appendix for details).

Therefore, we draw inspiration from these discrepancies and provide three key design principles. \textbf{\textit{i)} Detailed text description and image pairing}: The text must be well aligned with its corresponding image, providing information for both the foreground and background objects. \textbf{\textit{ii)} Reasonable object layout and background generation}: To avoid images looking like artificial Cut-Mixes and leverage the pre-existing information of LAION datasets, we must ensure that objects only co-exist in one image when it is possible to capture them in real-life, and that their position in the image makes sense. \textbf{\textit{iii)} High resolution}: This would ensure that our dataset caters to our end goal of generating high-quality, multi-concept personalized images.

\subsection{\textbf{\pipelinename} Pipeline}

Figure \ref{fig:data-creation} illustrates our \pipelinename creation pipeline. It comprises three main stages: 1) Object association and foreground segmentation, 2) LLM-guided object composition, 3) Background repainting and image recaptioning. While it would be optimal for the entire data generation process to be completely automated, the current state-of-the-art models~\cite{qin2022,lian2023llmgrounded,podell2023sdxl} still contain artifacts in every step. As our main goal is to provide a holistic benchmark in understanding current models' capabilities under complex image generation, the dataset preparation involves human-in-the-loop for intermediate and final cleaning.

\noindent\textbf{1) Object Association and Foreground Segmentation.} Our dataset begins with a set of $k$ objects $O = \{o_i\}_{i=1}^{k}$, where each object $o_i$ is represented by a set of $n$ images $X_{o_i} = \{x_j\}_{j=1}^n$. These sets are obtained from the datasets of DreamBooth, Custom Diffusion, and online copyright-free sources. We first find a subset of object combinations  $O' = \{o_a, o_b, ...\}, O' \in O$ that are intuitively likely to co-exist in a natural scene (e.g., dog, cat, and houseplant as depicted in Figure \ref{fig:data-creation}). 

We then grab one image from each of the source image sets representing objects within $O'$, forming the image set $X' = \{x_a \in X_{o_a}, x_b \in X_{o_b}, ...\}$ and apply DIS~\cite{qin2022} to obtain foreground for each image. We refer to these images as $\mathcal{D}(X')$ and their corresponding masks as $\mathcal{M}(\mathcal{D}(X'))$.  Note that DIS is a category-free saliency object detector, and thus is agnostic to the set of objects we are using. Interestingly, many of these objects that often co-exist are also the ones that fail on Custom Diffusion and even stable diffusion due to the similarity in their latent space. This conveniently makes our dataset more challenging and thus serves as a better benchmark for this task.

\noindent\textbf{2) LLM-Guided Object Composition.} We exploit LLM's zero-shot capability~\cite{devlin2018bert,brown2020language,openai2023gpt4,touvron2023llama} and inquire a probable set of bounding boxes given these sets of objects~\cite{lian2023llmgrounded}. Specifically, we show very few samples to ChatGPT explaining the task of providing bounding box points given object bounding boxes in the COCO dataset~\cite{lin2014microsoft} (template provided in Appendix), then ask ChatGPT to provide the set of bounding boxes given $O'$. We then place the individual images within $\mathcal{D}(X')$ following the bounding box location and given size to obtain a foreground image ready to be inpainted. We refer to this composited foreground image and its corresponding mask as $I_{fg}$ and $\mathcal{M}(I_{fg})$. We also obtain a set of prompts $P$ describing possible scenes $I_{fg}$ may be placed in by verifying its validness against the same LLM model (e.g., \textit{``in a garden"} is a valid prompt for dog, cat, and houseplant). 

The above-described method would occasionally lead to scaling problem, where some objects are unrealistically bigger than others (e.g., sheep is bigger than car). To alleviate this issue, we utilize the logical enhancements via GPT-4~\cite{openai2023gpt4} and request the scales for each bounding box to be realistic. Specifically, we prompt GPT-4 with the following: \texttt{[Given a list of object names, your task is to generate a reasonable scale ratio for these objects in real-world terms, where the ratio for the largest object is set to 1.0]}. These scales are subsequently used to adjust generate layouts to properly reflect real-world proportions.

\begin{figure*}[!t]
    \centering
    \includegraphics[width=\linewidth]{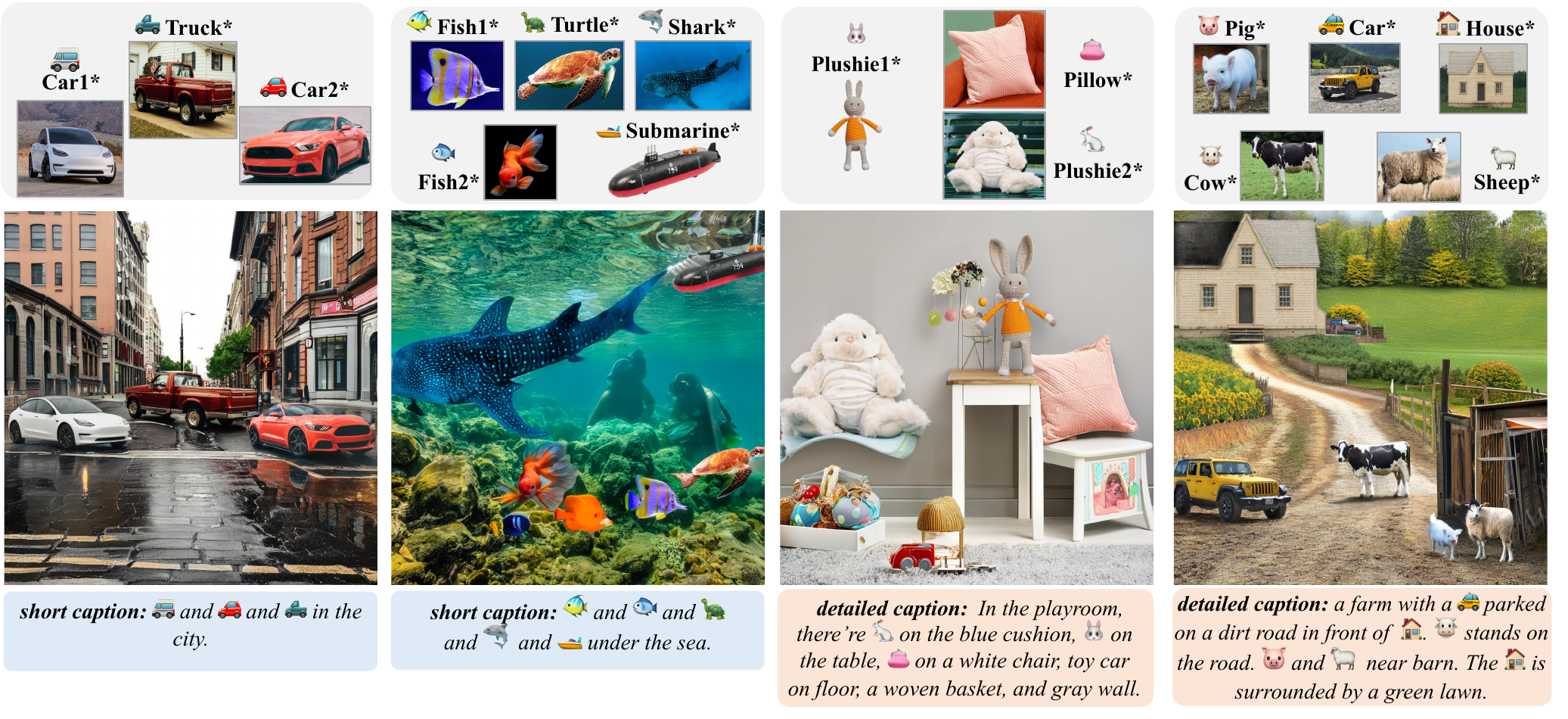}
    \vspace{-10pt}
    \caption{{\textbf{Examples of our \datasetname Dataset.} Our semi-automated generated dataset contains multiple personalized objects in complex compositions with high resolution, realistic images along with accurate text descriptions (short and detailed).
    } }
    \label{fig:data-example}
    \vspace{-5pt}
\end{figure*}

\noindent\textbf{3) Background Repainting and Image Recaptioning.} The most straightforward way to generate a background from $\mathcal{D}(X')$ and $\mathcal{M}(\mathcal{D}(X'))$ is to apply state-of-the-art text-to-image diffusion models to inpaint the background. However, we realize that forcing the model to generate a reasonable background (\textit{i.e.,} where objects do not look like cut-and-pasted) following a very vague text-prompt prior would often lead to unpredictable results. To simplify the problem setting, we realize that using a high resolution image reflecting the prompt and then ``repainting'' from it significantly improves the generation quality (detailed qualitative ablations are in the Appendix). Therefore, given a text-to-image diffusion inpainting model $f$ (in our case we use Stable-Diffusion-XL~\cite{podell2023sdxl}), we find a starting background image $I_{bg}$ from copyright-free sources\footnote{\url{https://unsplash.com/}} with a prompt $p \in P$ we are aiming to paint $I_{fg}$, then obtain the final image: $I_{O'} = f(I_{fg}, \mathcal{M}(I_{fg}), I_{bg})$. Note that during the repainting stage, we observe that utilizing a smoothed soft mask enhances the integration of the foreground object into the background as compared to a binary hard mask, and therefore we perform an average smoothing on $\mathcal{M}(I_{fg})$ with a $5\times 5$ window. The corresponding prompt to $I_{O'}$ is now a prompt listing out every object within $O'$ combined with $p$. 

As we strive to construct a holistic benchmark dataset, we enrich the diversity of text descriptions while adhering to the guidelines outlined in Section~\ref{sec:laion} and ensure that the text follows the image closely even in cases of extended length. Therefore, in light of the recent achievements of Multimodal Large Language Models (MLLMs)~\cite{dai2023instructblip,openai2023gpt4,zhu2023minigpt,liu2023visual}, we feed some of our final images into the LLaVA-1.5~\cite{liu2023visual} for automatic captioning using the specific instruction: \textit{"Describe what you see in this image in detail. The number of words is limited to 30"}. We constrain the word limit to accommodate CLIP's context constraints~\cite{radford2021learning}, which allow a maximum of 77 tokens. We highlight that our recaptioning is applied to ten object combinations $O'$ within the \datasetname dataset.

We repeat steps 1) to 3) to obtain a set of images and text descriptions per composition $O'$, and this is included in our final \datasetname dataset. Examples are shown in Figure~\ref{fig:data-example}.

\subsection{Dataset Statistics}

\begin{figure}[!t]
    \centering
    \includegraphics[width=\linewidth]{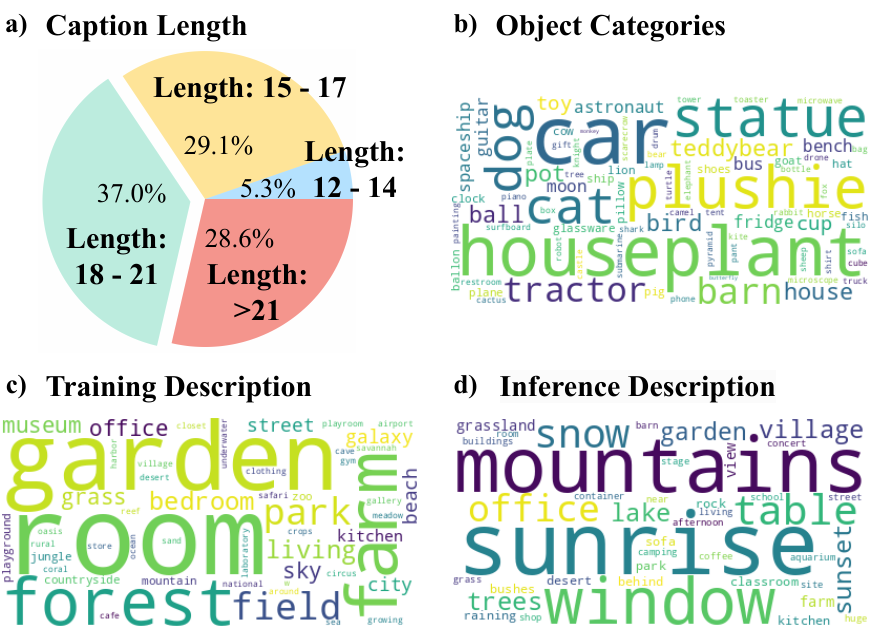}
    \vspace{-10pt}
    \caption{{\textbf{\datasetname Dataset Statistics.} a) A pie chart depicts that roughly 30\% of the images in \datasetname are paired with text descriptions over a length of 20 words. b) Word cloud of the categories used within the images to show the variety of objects used. c) and d) Word cloud of the frequent descriptions used during training and inference, which are very different to ensure fairness in comparison.
    } }
    \label{fig:dataset_stats}
    \vspace{-5pt}
\end{figure}

For \datasetname dataset, we collect 150 objects (some with a single image and others with multiple), and create 41 probable compositions (i.e., we refer to composition as a set $O'$) and over 10K images, then filter it manually down to 2684 images of best quality in terms of repainting results (more details in Appendix).

Figure \ref{fig:dataset_stats} presents statistics of our \datasetname. \textbf{a)} represents the composition of the number of words per caption (excluding the rare tokens we aim to learn and additional prompting strategies listed in Section \ref{sec:prompt}). Our average word length is 17.7 with approximately 30\% of the lengths extending beyond 20, where every word is specific and tailored to the image. On the other hand, \textbf{b)} represents the wide range of objects presented in our dataset, surpassing both CustomConcept101 and DreamBooth datasets.

The word cloud also presents various objects and background prompts we utilized during \textbf{c)} training and \textbf{d)} inference. Compared with previous benchmarks like DreamBooth and Custom Diffusion, our dataset covers a larger variety of objects with multi-concept combinations and therefore is a more comprehensive dataset for measuring the task of personalization.

\begin{figure*}[!t]
    \centering
    \includegraphics[width=\linewidth]{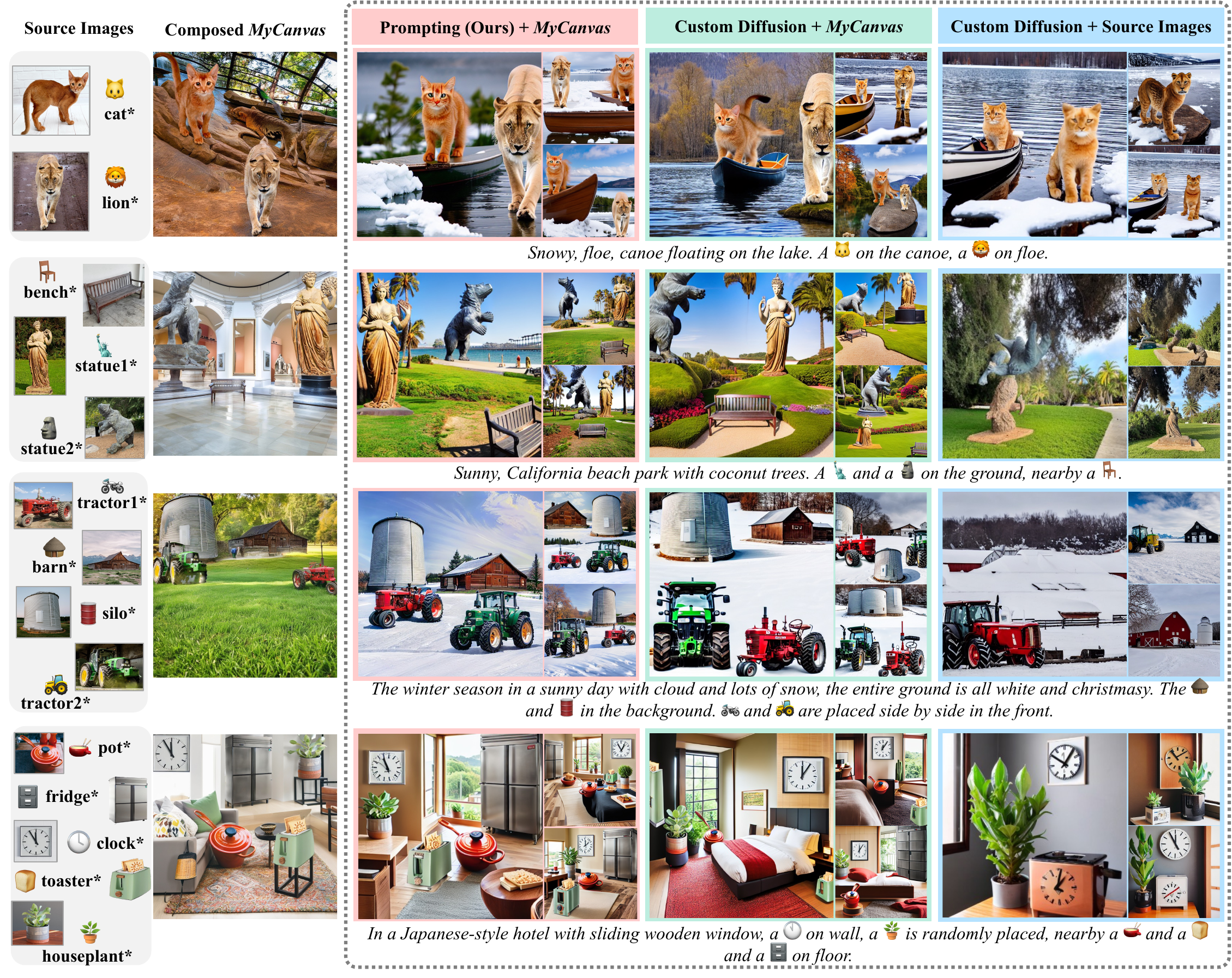}
    \vspace{-10pt}
    \caption{\textbf{Qualitative Results for Multi-Concept Composition. } We present four sets of results in ascending order of composition difficulty (more personalized concepts). Given training methods like Custom Diffusion, our \datasetname brings drastic improvements in disentangling object identities similar in the latent space (e.g., cat and lion, tractor1 and tractor2), preserving the distinctiveness of each object. Adding our prompting strategy gains even more improvements in aligning the caption during image generation (i.e., all the objects are properly reflected). More results are presented in the Appendix.
    } 
    \label{fig:results_complex}
    \vspace{-5pt}
\end{figure*}

\subsection{Improving Training-Time Text Prompts} \label{sec:prompt}

On top of designing a well-aligned prompt with the images within the dataset, we also take a step further in exploring what the best prompt design is during training. We share some of the empirical findings and its intuitions below:

\noindent\textbf{Global Composition Token.} 
Previous arts like DreamBooth have shown that they can learn to map a new token to very difficult, challenging concepts (e.g., an abstract style like Monet art). We adapt this concept to complex compositions. By introducing a global token alongside individual tokens for each object, our model gains enhanced capabilities in describing detailed scene arrangements, leading to more realistic and coherent image generation.

\noindent\textbf{Repeat Concept Token Prompts During Training.} We notice in a lot of cases where a complex composition involving multiple concepts could often lead to one or two concepts missing~\cite{yu2022scaling,chefer2023attend}. This could be due to the model sometimes forgetting the details given a very long prompt. Thus, we employ a strategy of repeating concept token prompts during training. This encourages the model to ensure the presence of each specified concept in the generated images, enhancing overall object persistence and completeness. 

\noindent\textbf{Incorporating Background Prompts.} 
We observe an issue where backgrounds are inadvertently learned with the object identity in the token feature space. As an effort to disentangle background and concept compositions, we make sure that background has to be stated within the training prompt to encourage concept tokens learning only the object identity.

\subsection{Personalized Composition Metric}
 As we increase the difficulty of the personalization challenge by increasing the number of objects, we realize an inherent tradeoff between the model not learning how to generate crucial details and the overfitting losing the capability of generating new backgrounds.
 The issue of this tradeoff is not reflected in previously used benchmarks as 1) no complex datasets like \datasetname were used for evaluation and 2) one can completely overfit to the training set and gain high quality results.

To overcome this, we draw inspiration from~\cite{bakr2023hrs,huang2023t2icompbench} and suggest two metrics. The first metric, Composition-Personalization-CLIP score (CP-CLIP), assesses the accuracy of composition and personalization. The second metric, Text-Image Alignment CLIP score (TI-CLIP), serves as an indicator of potential overfitting by evaluating the model's generalization quality across various textual backgrounds.

\noindent\textbf{Scene Composition and Personalization Accuracy.}
Different from existing benchmarks and metrics that primarily focus on the composition of general concepts~\cite{bakr2023hrs,huang2023t2icompbench}, our metric addresses two key questions: 1) \textit{Is each personalized concept mentioned in the text reflected during image generation? \textbf{(composition accuracy)}} and 2) \textit{Does the generated personalized concepts look similar to their source counterparts? \textbf{(fidelity)}}

To automate the full evaluation framework, we begin with a state-of-the-art model for open-vocabulary object detection, OWL-ViT~\cite{minderer2022simple}. The choice of open-vocabulary allows the capture of any objects within \datasetname dataset. Specifically, given a generated image $I_{gen}$ aiming to contain all objects within $O'$, we obtain a set of arbitrary cropped images specified by the predicted bounding boxes:

\begin{equation}
    B_{pred} = \{b_{pred_1}, b_{pred_2}, ...\} = \mbox{OWL}(I_{gen}, l_{O'}),
    \label{eqn:owl}
\end{equation}

where $l_{O'}$ are the individual labels within $O'$ which we use as the target vocabularies for OWL-ViT.

For every cropped image $b_{pred_i} \in B_{pred}$ we obtained from Eq. \ref{eqn:owl}, we compute an average clip score $S_{i,j}$ against the image set $X_{o_j}, o_j \in O'$ as the following:

\begin{equation}
    S_{i,j} = \frac{\sum_{x\in X_{o_j}}C(b_{pred_i}, x)}{|X_{o_j}|},
    \label{eqn:individual_clip}
\end{equation}

where $C(\cdot)$ computes the dot product between two normalized image features. The final personalization clip score for $b_{pred_i}$ is then:
\begin{equation}
    S_i = \max(\{S_{i,j}\}_{o_j \in O'}).
    \label{eqn:best_clip}
\end{equation}

If there is more than one bounding box corresponding to the same $o_j$, we remove all except the one with the highest score from $B_{pred}$ so the size $|B_{pred}|$ properly reflects how many personalized objects prompted by the text is reflected in the generated image.

Finally, we obtain an overall CP-CLIP score per image:

\begin{equation}
    \mbox{CP-CLIP}_{pred} = \frac{\sum_{b_{pred_i} \in B_{pred}}S_i}{|O'|}.
    \label{eqn:CP-Clip_Score}
\end{equation}
Note that the denominator is the number of objects within $O'$ and not the number of bounding boxes; this acts as a penalty when a particular personalized object is not reflected in the image $I_{gen}$. We do not penalize when there are more bounding boxes than intended, as the generative model should be able to freely generate more objects than requested as long as it follows the text guidance.

\noindent\textbf{Text-Image Alignment.}
To measure the amount of overfitting quantitatively, we calculate the TI-CLIP as a CLIP score between $I_{gen}$ and the prompt $p_{gen}$ that was used to generate $I_{gen}$. Note that while the formulation of TI-CLIP is very similar to CP-CLIP (\textit{i.e.,} one may think of TI-CLIP as a special case of the personalization clip score with the bounding box of the entire image and target of personalization being the text), it is evaluating an orthogonal concept of model's generalization quality and should thus be measured as a separate metric.

On a high level view, TI-CLIP measures the background prompt (without the objects) with the whole generated image; there is no reason to believe that the background is improved during personalization, so a maintenance in TI-CLIP should be what we are aiming for when increasing the CP-CLIP score. This shows that the model is not overfitting to training set backgrounds.

\noindent\textbf{Score Interpretability.} In practice, we realize that a good score in CP-CLIP is around 0.5, and TI-CLIP should be a score that maintains and not increases. We dive into the details of score interpretability in the Appendix.

\section{Experiments}

\begin{table}[t!]
    \centering
        \resizebox{\linewidth}{!}{
    \begin{tabular}{lcc|cc|cc}
    \toprule
    &\multicolumn{2}{c|}{ $<=$ 3 Objects} &\multicolumn{2}{c|}{4 Objects} &\multicolumn{2}{c}{5 Objects}\\
    & \textbf{CP-CLIP$\uparrow$}& \textbf{TI-CLIP}&\textbf{CP-CLIP$\uparrow$}&\textbf{TI-CLIP}&\textbf{CP-CLIP$\uparrow$}&\textbf{TI-CLIP}\\
     \cmidrule(lr){2-3} \cmidrule(lr){4-5} \cmidrule(lr){6-7} 

            CD + Source Images  &0.26& 0.16&0.21 & 0.13&0.23 & 0.17\\

            CD + \datasetname  & 0.41& 0.17& 0.47& 0.17& 0.50& 0.15\\

            Ours + \datasetname  & \cellcolor{LightGreen}\textbf{0.51}& 0.17& \cellcolor{LightGreen}\textbf{ 0.55}& 0.16&\cellcolor{LightGreen}\textbf{0.57 }& 0.14\\

    \bottomrule
    \end{tabular}
    }
    \vspace{-5pt}
    \caption{\textbf{Comparisons on Personalization Effectiveness}. Using our metric (CP-CLIP and TI-CLIP), we measure the quantitative performances under three different settings. CD denotes Custom Diffusion, and Prompting denotes our prompting strategy built on a CD training pipeline. Best score for CP-CLIP is highlighted in \textbf{bold}. Note that TI-CLIP acts as an indicator of whether we still reflect the prompt and thus should be a score that maintains \textit{roughly similar across all benchmarks}.}
    \label{tab:quantitative_analysis}
    \end{table}

\subsection{Baselines and Implementation Details.} We begin with the Custom Diffusion implementation and measure the quantitative and qualitative performance of 1) Custom Diffusion using the individual source concept images, 2) Custom Diffusion with composed \datasetname, and 3) our prompting strategy based on Custom Diffusion with \datasetname. We opt for Custom Diffusion due to its reproducible code base and extensive comparisons against prior methods. For each composition, we train a model for each abovementioned methods (training details in Appendix).

For evaluation, we use the best checkpoints for every composition. We choose a prompt per composition that is distinctively different from what is used during training. This allows us to better analyze the generalization capability of each model, as the background descriptions are unseen for all data. We use ViT-B-32 for as the backbone for both OWL-ViT bounding box extraction and calculating the two CLIP-based scores, CP-CLIP and TI-CLIP.

\subsection{Quantitative Analysis}
Table \ref{tab:quantitative_analysis} presents the outcomes across all compositions, organized by the number of objects. We use 41 text prompts, with 6 samples per prompt for each composition, resulting in a total of 246 generated images. It is evident that Custom Diffusion, when learning with the original source images, exhibits a 50\% decrease in performance compared to its counterpart utilizing our composed \datasetname dataset. By applying our prompting strategy to Custom Diffusion further amplifies the CP-CLIP score. Notably, our TI-CLIP score, indicative of background generalization, maintains consistency across all methods, ensuring that the observed increase in composition accuracy is not a consequence of overfitting.

\subsection{Qualitative Comparisons}
In Figure \ref{fig:results_complex}, we focus specifically on the qualitative outcomes on challenging and detailed prompts during inference. These prompts are carefully designed to test the model's ability to generate concepts in a novel scene distinct from training scenes, composing concepts with other known objects (e.g., \textit{a cat on the canoe, a lion on floe}), and describing the relative positions of concepts (e.g., \textit{side by side, in the background}). We compare qualitative results under three settings: 1) Custom Diffusion with the original source images, 2) Custom Diffusion with our \datasetname, composed dataset from the source, and 3) our prompting strategy built on top of Custom Diffusion with \datasetname. As depicted in Figure \ref{fig:results_complex}, even with highly challenging background descriptions, our composition strategy successfully disentangles objects that are similar in the latent space (e.g., lion and cat, two tractors), which are often failure cases in stable diffusion~\cite{rombach2022high}. Moreover, as the difficulty of the composition increases (i.e., descending each row increases the number of objects in the composition), our prompting method ensures that no concepts are left behind during the generation process. Notably, we demonstrate that by using \datasetname dataset, the generation quality of existing personalization models (e.g., Custom Diffusion) can be significantly enhanced (results for other methods can be found in Appendix).

\subsection{Ablation Study} 

\noindent\textbf{Evaluation of \datasetname Generation Quality.}
We developed a filtering tool (described in Appendix) to assess the quality of 800 images generated by our \pipelinename pipeline. We evaluate each image based on: \textit{1) the inclusion of personalized concepts, 2) their appropriate placement, and 3) the exclusion of visual artifacts,}
% Users rated each image on a scale of 1 to 5, and we tabulated the score distribution.
ranking them from 1 to 5. Subsequently, we aggregate these rankings to analyze the score distribution.
Only images rated 4/5 were added to the \datasetname dataset. Our findings in Table~\ref{tab:mycanvas-eval} indicate that generating high-quality images becomes more feasible with fewer than four concepts involved.

\begin{table}[h]
\vspace{-1pt}
\centering
\resizebox{0.99\columnwidth}{!}{
\begin{tabular}{c c c c c c c}
\toprule

 & \textbf{Rank: 1} & \textbf{Rank: 2} & \textbf{Rank: 3} & \textbf{Rank: 4} & \textbf{Rank: 5} & \textbf{Total Images} \\
\midrule
\textbf{$<$= 3 Concepts} &
   \textbf{9} (3.4 $\%$) & 
   \textbf{43} (16.3 $\%$) & 
   \textbf{72} (27.3 $\%$) & 
   \cellcolor{LightGreen}\textbf{84} (31.8 $\%$) & 
   \cellcolor{LightGreen}\textbf{56} (21.2 $\%$)& 
   \textbf{264}  \\ 

\midrule
\textbf{4 Concepts} &
   \textbf{16} (6.0 $\%$) & 
   \textbf{53} (19.8 $\%$) & 
   \textbf{112} (42.0 $\%$) & 
   \cellcolor{LightGreen}\textbf{54} (20.2 $\%$) & 
   \cellcolor{LightGreen}\textbf{32} (12.0 $\%$)& 
   \textbf{267}   \\ 

\textbf{5 Concepts} &
   \textbf{19} (7.1 $\%$) & 
   \textbf{63} (23.4 $\%$) & 
   \textbf{127} (47.2 $\%$) & 
   \cellcolor{LightGreen}\textbf{42} (15.6 $\%$) & 
   \cellcolor{LightGreen}\textbf{18} (6.7 $\%$)& 
   \textbf{269}   \\ 
\bottomrule

\end{tabular}
}
\vspace{-1pt}
\caption{
\textbf{Quality Evaluation of \datasetname (Rank: 1 to 5).} Our evaluation criteria include: \textit{1) inclusion of personalized concepts, 2) accuracy of their placement, and 3) absence of visual artifacts.} Images are ranked ranging from 1 to 5 based on these factors.
}
\label{tab:mycanvas-eval}

\end{table}

\begin{figure}[h]
    \vspace{-10pt}
    \centering
    \includegraphics[width=0.85\columnwidth]{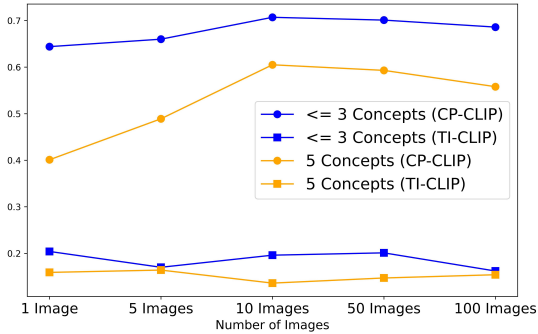}
    \caption{ \textbf{Training Performance Based on Dataset Size.} Training for compositions with $\leq 3$ concepts requires fewer images for efficacy, while stable performance is only achieved with 10 to 50 images for more than 4 concepts.
    } 
    \label{fig:train_data_size}
    \vspace{-12pt}
\end{figure}

\noindent\textbf{Training Data Size v.s. Number of Concepts.} We provide an analysis illustrated in Figure~\ref{fig:train_data_size}, training with varying number of images (1 to 100). While it is sufficient with very few image when training the compositions for $\leq 3$ concepts, the training stabilizes between 10 to 50 images when there are more than 4 concepts. This shows that our dataset size is more than enough to obtain stable performance. 

\begin{figure}[!t]
    \centering
    \includegraphics[width=\columnwidth]{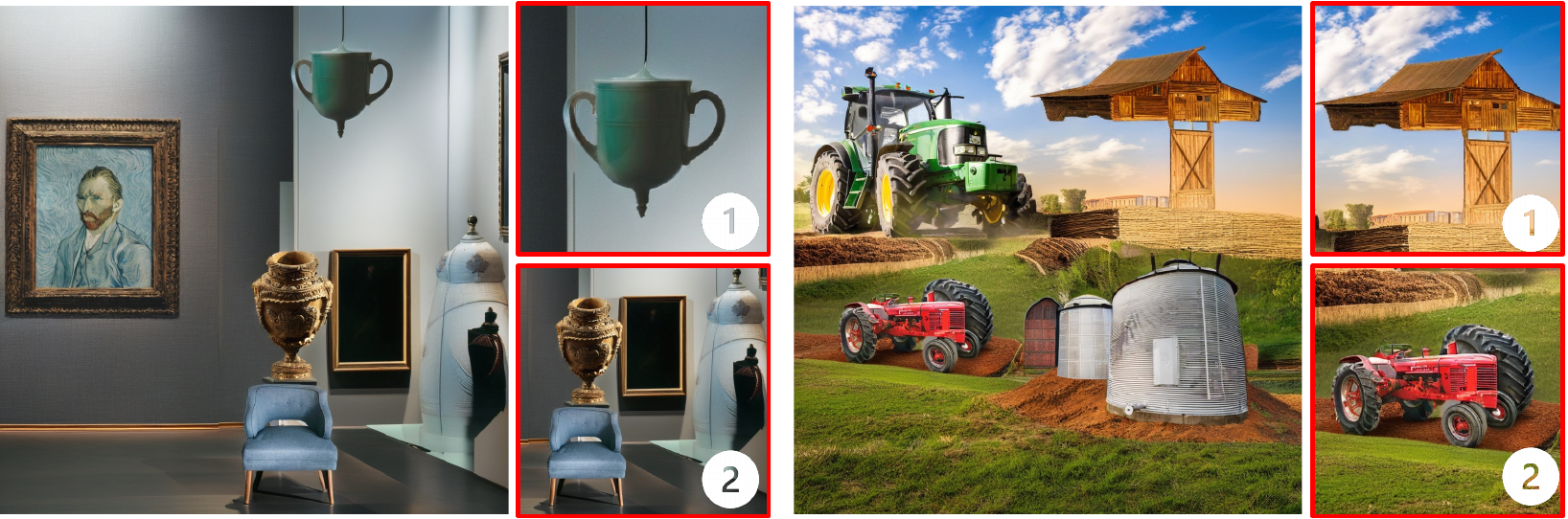}
    \vspace{-10pt}
    \caption{\textbf{Failure Dataset Creation Cases.} Our data creation pipeline fails in challenging composition scenarios due to two primary reasons: 1) If the LLM suggests unrealistic object positions, the object's identity may be altered during the diffusion inpainting process (e.g., a cup transforms into a pendant lamp), 2) diffusion inpainting occasionally introduces artifacts to the objects.
    } 
    \label{fig:limitations}
\end{figure}

\section{Conclusion}
In conclusion, we present \datasetname, a dataset with well-aligned image and text descriptions, as a benchmark for multi-concept personalization. We present extensive studies on our dataset, along with some training prompt amendments and a holistic metric, to show that improving data quality can lead to significantly better image generation for complex compositions.
We hope that our contributions serve as a foresight to the possibilities of personalized text-to-image generation and automated dataset creation.

\noindent\textbf{Limitations.} As depicted in Figure~\ref{fig:limitations}, our current data creation pipeline still contains defects, particularly in challenging scenarios. These challenges stem from the LLM offering impractical guidance on object positions, and the diffusion inpainting introducing artifacts to objects. For now, we resort to a semi-automated screening process to address these issues. Future work could focus on automating the filtering process and assessing dataset quality. In addition, with the new MLLMs having rich multi-modal understanding~\cite{openai2023gpt4,zhu2023minigpt}, we can include additional visual guidances for better bounding box generation.

{
    \small
    \bibliographystyle{ieeenat_fullname}
    \bibliography{main}

\begin{thebibliography}{44}
\providecommand{\natexlab}[1]{#1}
\providecommand{\url}[1]{\texttt{#1}}
\expandafter\ifx\csname urlstyle\endcsname\relax
  \providecommand{\doi}[1]{doi: #1}\else
  \providecommand{\doi}{doi: \begingroup \urlstyle{rm}\Url}\fi

\bibitem[Alaluf et~al.(2023)Alaluf, Richardson, Metzer, and Cohen-Or]{alaluf2023neural}
Yuval Alaluf, Elad Richardson, Gal Metzer, and Daniel Cohen-Or.
\newblock A neural space-time representation for text-to-image personalization.
\newblock \emph{arXiv preprint arXiv:2305.15391}, 2023.

\bibitem[Avrahami et~al.(2023)Avrahami, Aberman, Fried, Cohen-Or, and Lischinski]{avrahami2023break}
Omri Avrahami, Kfir Aberman, Ohad Fried, Daniel Cohen-Or, and Dani Lischinski.
\newblock Break-a-scene: Extracting multiple concepts from a single image.
\newblock \emph{arXiv preprint arXiv:2305.16311}, 2023.

\bibitem[Bakr et~al.(2023)Bakr, Sun, Shen, Khan, Li, and Elhoseiny]{bakr2023hrs}
Eslam~Mohamed Bakr, Pengzhan Sun, Xiaogian Shen, Faizan~Farooq Khan, Li~Erran Li, and Mohamed Elhoseiny.
\newblock Hrs-bench: Holistic, reliable and scalable benchmark for text-to-image models.
\newblock In \emph{Proceedings of the IEEE/CVF International Conference on Computer Vision}, pages 20041--20053, 2023.

\bibitem[Betker et~al.(2023)Betker, Goh, Jing, Brooks, Wang, Lia, Ouyang, Zhuang, Lee, Guo, Manassra, Dhariwal, Chu, Jiao, and Ramesh]{dalle3}
James Betker, Gabriel Goh, Li Jing, Tim Brooks, Jianfeng Wang, Linjie Lia, Long Ouyang, Juntang Zhuang, Joyce Lee, Yufei Guo, Wesam Manassra, Prafulla Dhariwal, Casey Chu, Yunxin Jiao, and Aditya Ramesh.
\newblock Improving image generation with better captions.
\newblock \url{https://cdn.openai.com/papers/dall-e-3.pdf}, 2023.

\bibitem[Brown et~al.(2020)Brown, Mann, Ryder, Subbiah, Kaplan, Dhariwal, Neelakantan, Shyam, Sastry, Askell, et~al.]{brown2020language}
Tom Brown, Benjamin Mann, Nick Ryder, Melanie Subbiah, Jared~D Kaplan, Prafulla Dhariwal, Arvind Neelakantan, Pranav Shyam, Girish Sastry, Amanda Askell, et~al.
\newblock Language models are few-shot learners.
\newblock \emph{Advances in neural information processing systems}, 33:\penalty0 1877--1901, 2020.

\bibitem[Chefer et~al.(2023)Chefer, Alaluf, Vinker, Wolf, and Cohen-Or]{chefer2023attend}
Hila Chefer, Yuval Alaluf, Yael Vinker, Lior Wolf, and Daniel Cohen-Or.
\newblock Attend-and-excite: Attention-based semantic guidance for text-to-image diffusion models.
\newblock \emph{ACM Transactions on Graphics (TOG)}, 42\penalty0 (4):\penalty0 1--10, 2023.

\bibitem[Chen et~al.(2023{\natexlab{a}})Chen, Zhang, Wang, Duan, Zhou, and Zhu]{chen2023disenbooth}
Hong Chen, Yipeng Zhang, Xin Wang, Xuguang Duan, Yuwei Zhou, and Wenwu Zhu.
\newblock Disenbooth: Identity-preserving disentangled tuning for subject-driven text-to-image generation.
\newblock \emph{arXiv preprint arXiv:2305.03374}, 2023{\natexlab{a}}.

\bibitem[Chen et~al.(2023{\natexlab{b}})Chen, Zhao, Liu, Ding, Song, Wang, Wang, Yang, Liu, Du, et~al.]{chen2023photoverse}
Li Chen, Mengyi Zhao, Yiheng Liu, Mingxu Ding, Yangyang Song, Shizun Wang, Xu Wang, Hao Yang, Jing Liu, Kang Du, et~al.
\newblock Photoverse: Tuning-free image customization with text-to-image diffusion models.
\newblock \emph{arXiv preprint arXiv:2309.05793}, 2023{\natexlab{b}}.

\bibitem[Chen et~al.(2023{\natexlab{c}})Chen, Hu, Li, Ruiz, Jia, Chang, and Cohen]{chen2023suti}
Wenhu Chen, Hexiang Hu, Yandong Li, Nataniel Ruiz, Xuhui Jia, Ming-Wei Chang, and William~W Cohen.
\newblock Subject-driven text-to-image generation via apprenticeship learning.
\newblock \emph{arXiv preprint arXiv:2304.00186}, 2023{\natexlab{c}}.

\bibitem[Cho et~al.(2023)Cho, Zala, and Bansal]{cho2023dall}
Jaemin Cho, Abhay Zala, and Mohit Bansal.
\newblock Dall-eval: Probing the reasoning skills and social biases of text-to-image generation models.
\newblock In \emph{Proceedings of the IEEE/CVF International Conference on Computer Vision}, pages 3043--3054, 2023.

\bibitem[Dai et~al.(2023)Dai, Li, Li, Tiong, Zhao, Wang, Li, Fung, and Hoi]{dai2023instructblip}
Wenliang Dai, Junnan Li, Dongxu Li, Anthony Meng~Huat Tiong, Junqi Zhao, Weisheng Wang, Boyang Li, Pascale Fung, and Steven Hoi.
\newblock Instructblip: Towards general-purpose vision-language models with instruction tuning, 2023.

\bibitem[Devlin et~al.(2018)Devlin, Chang, Lee, and Toutanova]{devlin2018bert}
Jacob Devlin, Ming-Wei Chang, Kenton Lee, and Kristina Toutanova.
\newblock Bert: Pre-training of deep bidirectional transformers for language understanding.
\newblock \emph{arXiv preprint arXiv:1810.04805}, 2018.

\bibitem[Dinh et~al.(2022)Dinh, Nguyen, and Hua]{dinh2022tise}
Tan~M Dinh, Rang Nguyen, and Binh-Son Hua.
\newblock Tise: Bag of metrics for text-to-image synthesis evaluation.
\newblock In \emph{European Conference on Computer Vision}, pages 594--609. Springer, 2022.

\bibitem[Gal et~al.(2022)Gal, Alaluf, Atzmon, Patashnik, Bermano, Chechik, and Cohen-Or]{gal2022image}
Rinon Gal, Yuval Alaluf, Yuval Atzmon, Or Patashnik, Amit~H Bermano, Gal Chechik, and Daniel Cohen-Or.
\newblock An image is worth one word: Personalizing text-to-image generation using textual inversion.
\newblock \emph{arXiv preprint arXiv:2208.01618}, 2022.

\bibitem[Gu et~al.(2022)Gu, Chen, Bao, Wen, Zhang, Chen, Yuan, and Guo]{gu2022vector}
Shuyang Gu, Dong Chen, Jianmin Bao, Fang Wen, Bo Zhang, Dongdong Chen, Lu Yuan, and Baining Guo.
\newblock Vector quantized diffusion model for text-to-image synthesis, 2022.

\bibitem[Gu et~al.(2023)Gu, Wang, Wu, Shi, Chen, Fan, Xiao, Zhao, Chang, Wu, et~al.]{gu2023mix}
Yuchao Gu, Xintao Wang, Jay~Zhangjie Wu, Yujun Shi, Yunpeng Chen, Zihan Fan, Wuyou Xiao, Rui Zhao, Shuning Chang, Weijia Wu, et~al.
\newblock Mix-of-show: Decentralized low-rank adaptation for multi-concept customization of diffusion models.
\newblock \emph{arXiv preprint arXiv:2305.18292}, 2023.

\bibitem[Han et~al.(2023)Han, Li, Zhang, Milanfar, Metaxas, and Yang]{Han_2023_ICCV}
Ligong Han, Yinxiao Li, Han Zhang, Peyman Milanfar, Dimitris Metaxas, and Feng Yang.
\newblock Svdiff: Compact parameter space for diffusion fine-tuning.
\newblock In \emph{Proceedings of the IEEE/CVF International Conference on Computer Vision (ICCV)}, pages 7323--7334, 2023.

\bibitem[Huang et~al.(2023)Huang, Sun, Xie, Li, and Liu]{huang2023t2icompbench}
Kaiyi Huang, Kaiyue Sun, Enze Xie, Zhenguo Li, and Xihui Liu.
\newblock T2i-compbench: A comprehensive benchmark for open-world compositional text-to-image generation.
\newblock \emph{arXiv preprint arXiv:2307.06350}, 2023.

\bibitem[Kumari et~al.(2023)Kumari, Zhang, Zhang, Shechtman, and Zhu]{kumari2023multi}
Nupur Kumari, Bingliang Zhang, Richard Zhang, Eli Shechtman, and Jun-Yan Zhu.
\newblock Multi-concept customization of text-to-image diffusion.
\newblock In \emph{Proceedings of the IEEE/CVF Conference on Computer Vision and Pattern Recognition}, pages 1931--1941, 2023.

\bibitem[Lian et~al.(2023)Lian, Li, Yala, and Darrell]{lian2023llmgrounded}
Long Lian, Boyi Li, Adam Yala, and Trevor Darrell.
\newblock Llm-grounded diffusion: Enhancing prompt understanding of text-to-image diffusion models with large language models.
\newblock \emph{arXiv preprint arXiv:2305.13655}, 2023.

\bibitem[Lin et~al.(2014)Lin, Maire, Belongie, Hays, Perona, Ramanan, Doll{\'a}r, and Zitnick]{lin2014microsoft}
Tsung-Yi Lin, Michael Maire, Serge Belongie, James Hays, Pietro Perona, Deva Ramanan, Piotr Doll{\'a}r, and C~Lawrence Zitnick.
\newblock Microsoft coco: Common objects in context.
\newblock In \emph{Computer Vision--ECCV 2014: 13th European Conference, Zurich, Switzerland, September 6-12, 2014, Proceedings, Part V 13}, pages 740--755. Springer, 2014.

\bibitem[Liu et~al.(2023{\natexlab{a}})Liu, Li, Wu, and Lee]{liu2023visual}
Haotian Liu, Chunyuan Li, Qingyang Wu, and Yong~Jae Lee.
\newblock Visual instruction tuning.
\newblock \emph{arXiv preprint arXiv:2304.08485}, 2023{\natexlab{a}}.

\bibitem[Liu et~al.(2023{\natexlab{b}})Liu, Zhang, Shen, Zheng, Zhu, Feng, Liu, Zhao, Zhou, and Cao]{liu2023cones}
Zhiheng Liu, Yifei Zhang, Yujun Shen, Kecheng Zheng, Kai Zhu, Ruili Feng, Yu Liu, Deli Zhao, Jingren Zhou, and Yang Cao.
\newblock Cones 2: Customizable image synthesis with multiple subjects.
\newblock \emph{arXiv preprint arXiv:2305.19327}, 2023{\natexlab{b}}.

\bibitem[Ma et~al.(2023)Ma, Liang, Chen, and Lu]{ma2023subject}
Jian Ma, Junhao Liang, Chen Chen, and Haonan Lu.
\newblock Subject-diffusion: Open domain personalized text-to-image generation without test-time fine-tuning.
\newblock \emph{arXiv preprint arXiv:2307.11410}, 2023.

\bibitem[Minderer et~al.(2022)Minderer, Gritsenko, Stone, Neumann, Weissenborn, Dosovitskiy, Mahendran, Arnab, Dehghani, Shen, et~al.]{minderer2022simple}
Matthias Minderer, Alexey Gritsenko, Austin Stone, Maxim Neumann, Dirk Weissenborn, Alexey Dosovitskiy, Aravindh Mahendran, Anurag Arnab, Mostafa Dehghani, Zhuoran Shen, et~al.
\newblock Simple open-vocabulary object detection.
\newblock In \emph{European Conference on Computer Vision}, pages 728--755. Springer, 2022.

\bibitem[Nichol et~al.(2021)Nichol, Dhariwal, Ramesh, Shyam, Mishkin, McGrew, Sutskever, and Chen]{nichol2021glide}
Alex Nichol, Prafulla Dhariwal, Aditya Ramesh, Pranav Shyam, Pamela Mishkin, Bob McGrew, Ilya Sutskever, and Mark Chen.
\newblock Glide: Towards photorealistic image generation and editing with text-guided diffusion models.
\newblock \emph{arXiv preprint arXiv:2112.10741}, 2021.

\bibitem[OpenAI(2023)]{openai2023gpt4}
OpenAI.
\newblock Gpt-4 technical report, 2023.

\bibitem[Petsiuk et~al.(2022)Petsiuk, Siemenn, Surbehera, Chin, Tyser, Hunter, Raghavan, Hicke, Plummer, Kerret, et~al.]{petsiuk2022human}
Vitali Petsiuk, Alexander~E Siemenn, Saisamrit Surbehera, Zad Chin, Keith Tyser, Gregory Hunter, Arvind Raghavan, Yann Hicke, Bryan~A Plummer, Ori Kerret, et~al.
\newblock Human evaluation of text-to-image models on a multi-task benchmark.
\newblock \emph{arXiv preprint arXiv:2211.12112}, 2022.

\bibitem[Podell et~al.(2023)Podell, English, Lacey, Blattmann, Dockhorn, M{\"u}ller, Penna, and Rombach]{podell2023sdxl}
Dustin Podell, Zion English, Kyle Lacey, Andreas Blattmann, Tim Dockhorn, Jonas M{\"u}ller, Joe Penna, and Robin Rombach.
\newblock Sdxl: Improving latent diffusion models for high-resolution image synthesis.
\newblock \emph{arXiv preprint arXiv:2307.01952}, 2023.

\bibitem[Qin et~al.(2022)Qin, Dai, Hu, Fan, Shao, and Gool]{qin2022}
Xuebin Qin, Hang Dai, Xiaobin Hu, Deng-Ping Fan, Ling Shao, and Luc~Van Gool.
\newblock Highly accurate dichotomous image segmentation.
\newblock In \emph{ECCV}, 2022.

\bibitem[Radford et~al.(2021)Radford, Kim, Hallacy, Ramesh, Goh, Agarwal, Sastry, Askell, Mishkin, Clark, et~al.]{radford2021learning}
Alec Radford, Jong~Wook Kim, Chris Hallacy, Aditya Ramesh, Gabriel Goh, Sandhini Agarwal, Girish Sastry, Amanda Askell, Pamela Mishkin, Jack Clark, et~al.
\newblock Learning transferable visual models from natural language supervision.
\newblock In \emph{International conference on machine learning}, pages 8748--8763. PMLR, 2021.

\bibitem[Raffel et~al.(2020)Raffel, Shazeer, Roberts, Lee, Narang, Matena, Zhou, Li, and Liu]{raffel2020exploring}
Colin Raffel, Noam Shazeer, Adam Roberts, Katherine Lee, Sharan Narang, Michael Matena, Yanqi Zhou, Wei Li, and Peter~J Liu.
\newblock Exploring the limits of transfer learning with a unified text-to-text transformer.
\newblock \emph{The Journal of Machine Learning Research}, 21\penalty0 (1):\penalty0 5485--5551, 2020.

\bibitem[Ramesh et~al.(2022)Ramesh, Dhariwal, Nichol, Chu, and Chen]{ramesh2022hierarchical}
Aditya Ramesh, Prafulla Dhariwal, Alex Nichol, Casey Chu, and Mark Chen.
\newblock Hierarchical text-conditional image generation with clip latents, 2022.

\bibitem[Rombach et~al.(2022)Rombach, Blattmann, Lorenz, Esser, and Ommer]{rombach2022high}
Robin Rombach, Andreas Blattmann, Dominik Lorenz, Patrick Esser, and Bj{\"o}rn Ommer.
\newblock High-resolution image synthesis with latent diffusion models.
\newblock In \emph{Proceedings of the IEEE/CVF conference on computer vision and pattern recognition}, pages 10684--10695, 2022.

\bibitem[Ruiz et~al.(2023)Ruiz, Li, Jampani, Pritch, Rubinstein, and Aberman]{ruiz2023dreambooth}
Nataniel Ruiz, Yuanzhen Li, Varun Jampani, Yael Pritch, Michael Rubinstein, and Kfir Aberman.
\newblock Dreambooth: Fine tuning text-to-image diffusion models for subject-driven generation.
\newblock In \emph{Proceedings of the IEEE/CVF Conference on Computer Vision and Pattern Recognition}, pages 22500--22510, 2023.

\bibitem[Saharia et~al.(2022)Saharia, Chan, Saxena, Li, Whang, Denton, Ghasemipour, Gontijo~Lopes, Karagol~Ayan, Salimans, et~al.]{saharia2022photorealistic}
Chitwan Saharia, William Chan, Saurabh Saxena, Lala Li, Jay Whang, Emily~L Denton, Kamyar Ghasemipour, Raphael Gontijo~Lopes, Burcu Karagol~Ayan, Tim Salimans, et~al.
\newblock Photorealistic text-to-image diffusion models with deep language understanding.
\newblock \emph{Advances in Neural Information Processing Systems}, 35:\penalty0 36479--36494, 2022.

\bibitem[Schuhmann et~al.(2022)Schuhmann, Beaumont, Vencu, Gordon, Wightman, Cherti, Coombes, Katta, Mullis, Wortsman, et~al.]{schuhmann2022laion}
Christoph Schuhmann, Romain Beaumont, Richard Vencu, Cade Gordon, Ross Wightman, Mehdi Cherti, Theo Coombes, Aarush Katta, Clayton Mullis, Mitchell Wortsman, et~al.
\newblock Laion-5b: An open large-scale dataset for training next generation image-text models.
\newblock \emph{Advances in Neural Information Processing Systems}, 35:\penalty0 25278--25294, 2022.

\bibitem[Segalis et~al.(2023)Segalis, Valevski, Lumen, Matias, and Leviathan]{segalis2023picture}
Eyal Segalis, Dani Valevski, Danny Lumen, Yossi Matias, and Yaniv Leviathan.
\newblock A picture is worth a thousand words: Principled recaptioning improves image generation.
\newblock \emph{arXiv preprint arXiv:2310.16656}, 2023.

\bibitem[Touvron et~al.(2023)Touvron, Martin, Stone, Albert, Almahairi, Babaei, Bashlykov, Batra, Bhargava, Bhosale, et~al.]{touvron2023llama}
Hugo Touvron, Louis Martin, Kevin Stone, Peter Albert, Amjad Almahairi, Yasmine Babaei, Nikolay Bashlykov, Soumya Batra, Prajjwal Bhargava, Shruti Bhosale, et~al.
\newblock Llama 2: Open foundation and fine-tuned chat models.
\newblock \emph{arXiv preprint arXiv:2307.09288}, 2023.

\bibitem[Wei et~al.(2023)Wei, Zhang, Ji, Bai, Zhang, and Zuo]{Wei_2023_ICCV}
Yuxiang Wei, Yabo Zhang, Zhilong Ji, Jinfeng Bai, Lei Zhang, and Wangmeng Zuo.
\newblock Elite: Encoding visual concepts into textual embeddings for customized text-to-image generation.
\newblock In \emph{Proceedings of the IEEE/CVF International Conference on Computer Vision (ICCV)}, pages 15943--15953, 2023.

\bibitem[Wu et~al.(2023)Wu, Yu, Zhu, Wang, and Bai]{wu2023singleinsert}
Zijie Wu, Chaohui Yu, Zhen Zhu, Fan Wang, and Xiang Bai.
\newblock Singleinsert: Inserting new concepts from a single image into text-to-image models for flexible editing.
\newblock \emph{arXiv preprint arXiv:2310.08094}, 2023.

\bibitem[Xiao et~al.(2023)Xiao, Yin, Freeman, Durand, and Han]{xiao2023fastcomposer}
Guangxuan Xiao, Tianwei Yin, William~T Freeman, Fr{\'e}do Durand, and Song Han.
\newblock Fastcomposer: Tuning-free multi-subject image generation with localized attention.
\newblock \emph{arXiv preprint arXiv:2305.10431}, 2023.

\bibitem[Yu et~al.(2022)Yu, Xu, Koh, Luong, Baid, Wang, Vasudevan, Ku, Yang, Ayan, et~al.]{yu2022scaling}
Jiahui Yu, Yuanzhong Xu, Jing~Yu Koh, Thang Luong, Gunjan Baid, Zirui Wang, Vijay Vasudevan, Alexander Ku, Yinfei Yang, Burcu~Karagol Ayan, et~al.
\newblock Scaling autoregressive models for content-rich text-to-image generation.
\newblock \emph{arXiv preprint arXiv:2206.10789}, 2\penalty0 (3):\penalty0 5, 2022.

\bibitem[Zhu et~al.(2023)Zhu, Chen, Shen, Li, and Elhoseiny]{zhu2023minigpt}
Deyao Zhu, Jun Chen, Xiaoqian Shen, Xiang Li, and Mohamed Elhoseiny.
\newblock Minigpt-4: Enhancing vision-language understanding with advanced large language models.
\newblock \emph{arXiv preprint arXiv:2304.10592}, 2023.

\end{thebibliography}
}

% WARNING: do not forget to delete the supplementary pages from your submission 
\clearpage

\section*{Appendix}
In this supplementary, we will first provide additional details of the analysis on LAION dataset (Section~\ref{sec:supp-laion}). In Section~\ref{sec:supp-dataset}, we give additional details about the process of collecting our \datasetname dataset. Section~\ref{sec:supp-experiment} provides additional experiments about our \datasetname dataset used by DreamBooth. Section~\ref{sec:supp-metrics} discusses more details on metrics. Section~\ref{sec:supp-implement-details} provides additional implementation details of our approach.

\section{Laion Reflections}
\label{sec:supp-laion}

\begin{figure}[!t]
    \centering
    \includegraphics[width=1.\linewidth]{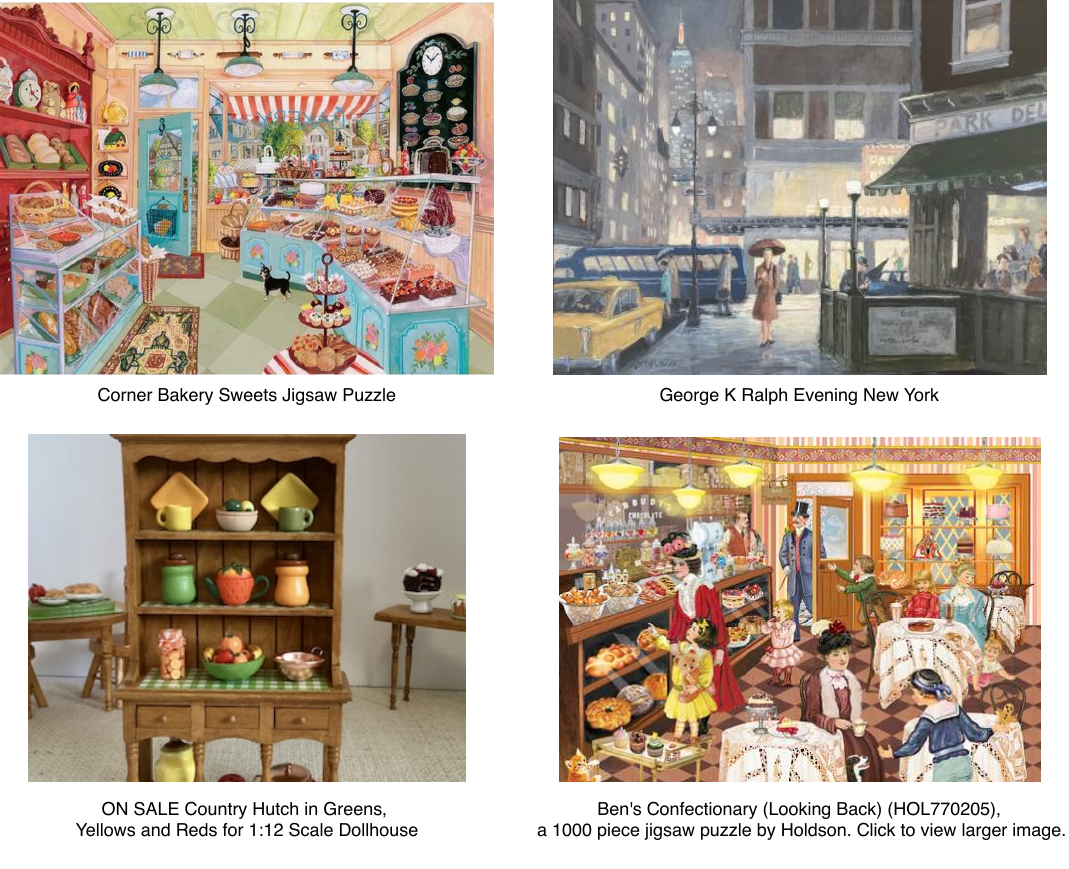}
    \vspace{-10pt}
    \caption{\textbf{Examples of Most Aesthetic LAION text-image pairs. } These examples indicate text-image misalignments within LAION dataset~\cite{schuhmann2022laion}, including instances where descriptions are either insufficient (e.g., "Bakery Sweets" lacking specific item descriptions, or "Evening New York" omitting details about lamps or cars) or include unnecessary elements (e.g., "ON SALE," "1:12," and "HOL770205"). Such discrepancies pose challenges during model fine-tuning for high-quality multi-concept image generation.
    } 
    \label{fig:sup_laion}
    \vspace{-5pt}
\end{figure}

\noindent\textbf{Analysis.} Figure \ref{fig:sup_laion} depicts the text and image pairs of the four most aesthetic images from LAION~\cite{schuhmann2022laion}. Specifically, they are of the highest aesthetic scores in the LAION-aesthetic dataset. Qualitative results show that images with high aesthetics are highly correlated with the complexity of the images.

It is important to note that there exist high misalignments within this dataset. These errors include the lack of descriptions (e.g., ``Bakery Sweets" without specific descriptions of the items within, or ``Evening New York" without describing the lamps or cars in the image), unnecessary descriptions (e.g.,``ON SALE", ``1:12" and ``HOL770205"), which could all sabotage the quality of multi-concept image generation when fine-tuning the model.

While our current data quality is small as we are targeting the task of personalization, we envision that future work could focus on improving the data quantity as well as quality to better improve foundation generative models.

\section{\datasetname Dataset}
\label{sec:supp-dataset}

\begin{figure*}[!t]
    \centering
    \includegraphics[width=0.80\linewidth]{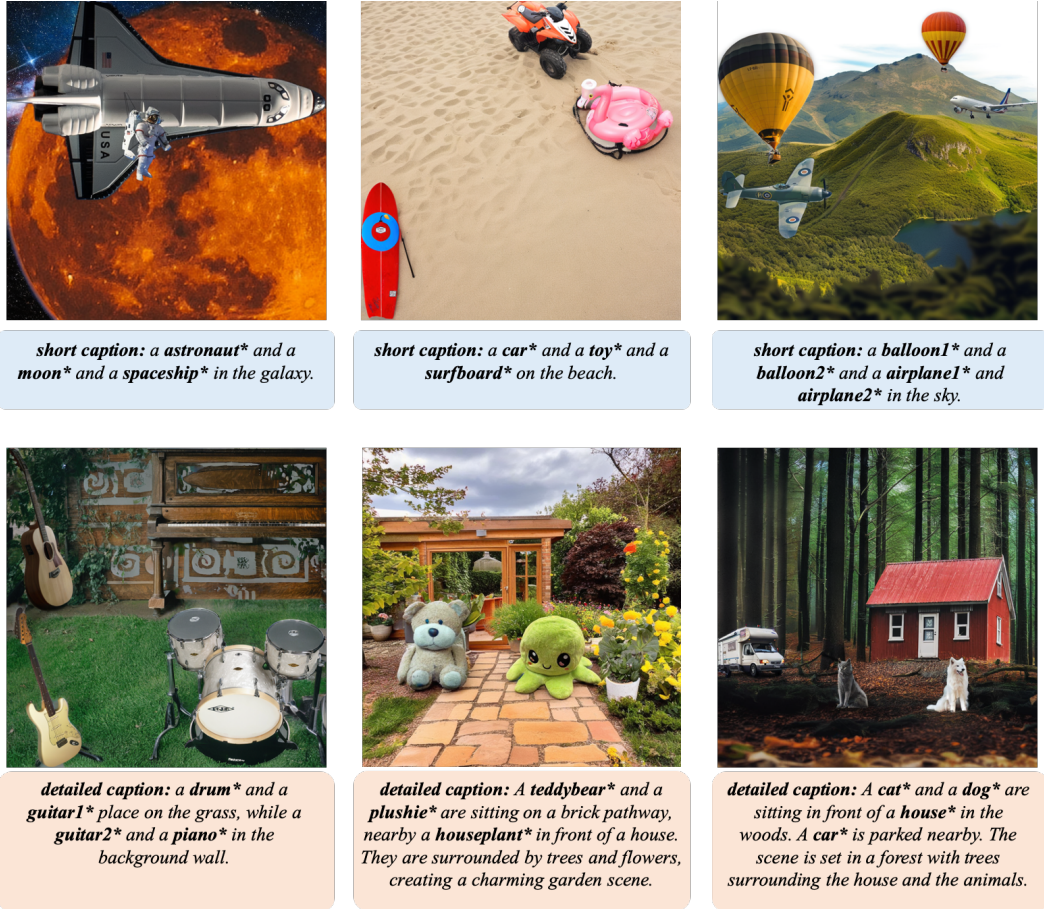}

    \caption{\textbf{Diverse Composition Examples in \datasetname. } We show examples from our \datasetname, featuring diverse compositions comprising 3 to 5 concepts set against distinct backgrounds. This collection aligns with accurate text descriptions.
    } 
    \label{fig:sup_data}

\end{figure*}

\noindent\textbf{\datasetname Examples.}
In Figure~\ref{fig:sup_data}, we show more examples within 41 compositions in \datasetname. These compositions encompass varying numbers of concepts set against distinct backgrounds, ranging from gardens to sky and the sea. Accompanying each composition are precise and fitting text descriptions, ensuring a comprehensive portrayal of the images, be it in succinct or detailed form.

\begin{figure*}[!t]
    \centering
    \includegraphics[width=\linewidth]{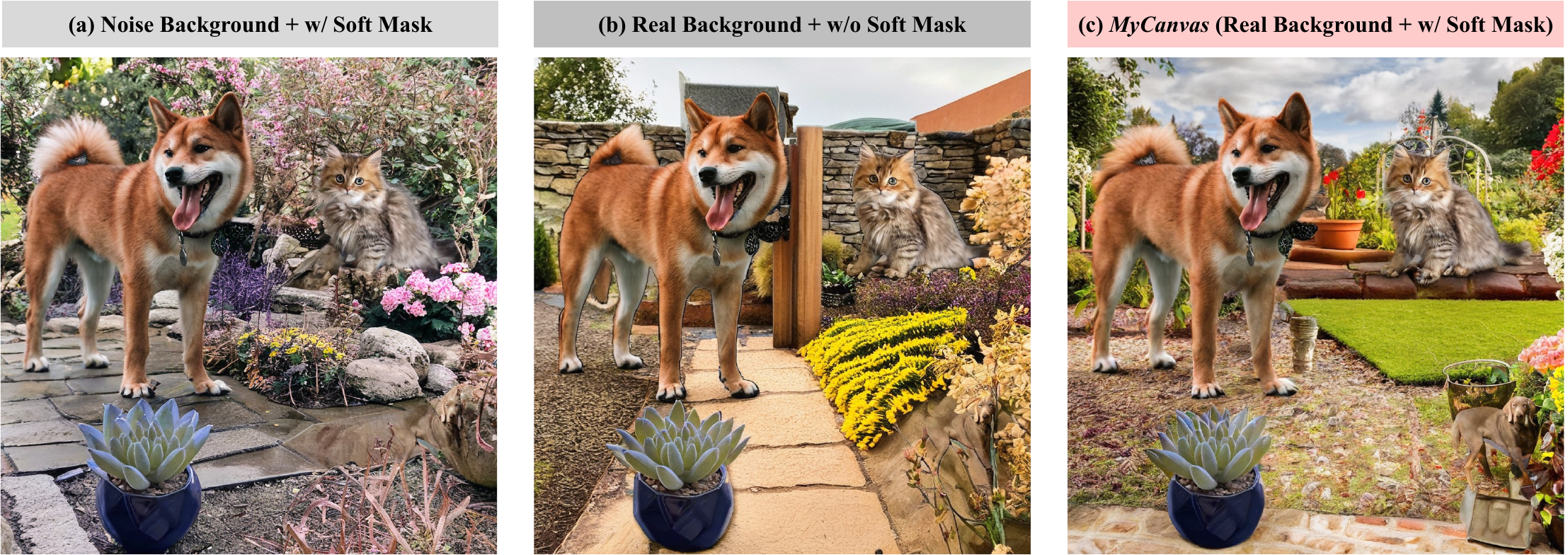}
    \vspace{-10pt}
    \caption{\textbf{Ablations of \datasetname Data Creation Pipeline. } In (c), we enhance image data quality by incorporating realistic background images and employing soft masks for foreground objects. Subfigure (a) illustrates that utilizing noise directly as the background for diffusion inpainting leads to the generation of unrealistic and non-colorful backgrounds. Meanwhile, (b) demonstrates the significance of employing soft masks on foreground objects to prevent the creation of visible boundary lines between objects and the background, avoiding the appearance of copy-paste-like artifacts.
    } 
    \label{fig:sup_data_ablation}

\end{figure*}

\noindent\textbf{Ablations on Background Repainting.} In Figure~\ref{fig:sup_data_ablation}, we conduct ablations on two key aspects: 1) comparing the initial noise background to a realistic image background, and 2) evaluating the impact of using a hard or soft mask on the foreground object. For the diffusion inpainting model $f(.)$~\cite{podell2023sdxl}, we provide a foreground image $I_{fg}$ with composed concepts $O'$ along with a mask $\mathcal{M}(I_{fg})$ specifying the concepts, while the model $f(.)$ is able to inpaint the background based on the given prompt $p$. 

In our experiment, we use a composed foreground image $I_{fg}$ where a black background is included. Initially, we attempt to enhance this by adding noise to the background and inputting it into the diffusion inpainting model to generate a background based on the provided background prompt (e.g., in the garden). However, Figure~\ref{fig:sup_data_ablation}-(a) reveals that allowing the diffusion model to generate the background from the scratch directly from the initial noise consistently results in unrealistic backgrounds and artifacts. 

Moreover, employing $\mathcal{M}(I_{fg})$ as the input to the diffusion inpainting model reveals visible boundary lines between objects and the background, resembling copy-paste-like artifacts (Figure~\ref{fig:sup_data_ablation}-(b)). To mitigate this, during the repainting stage, we observe use a smoothed soft mask to enhance the integration of the foreground object into the background. Therefore, we perform an average smoothing on $\mathcal{M}(I_{fg})$ with a 5 × 5 window. This section explores these ablation strategies and provides improvements into foreground enhancement techniques.

\begin{figure*}[!t]
    \centering
    \includegraphics[width=0.9\linewidth]{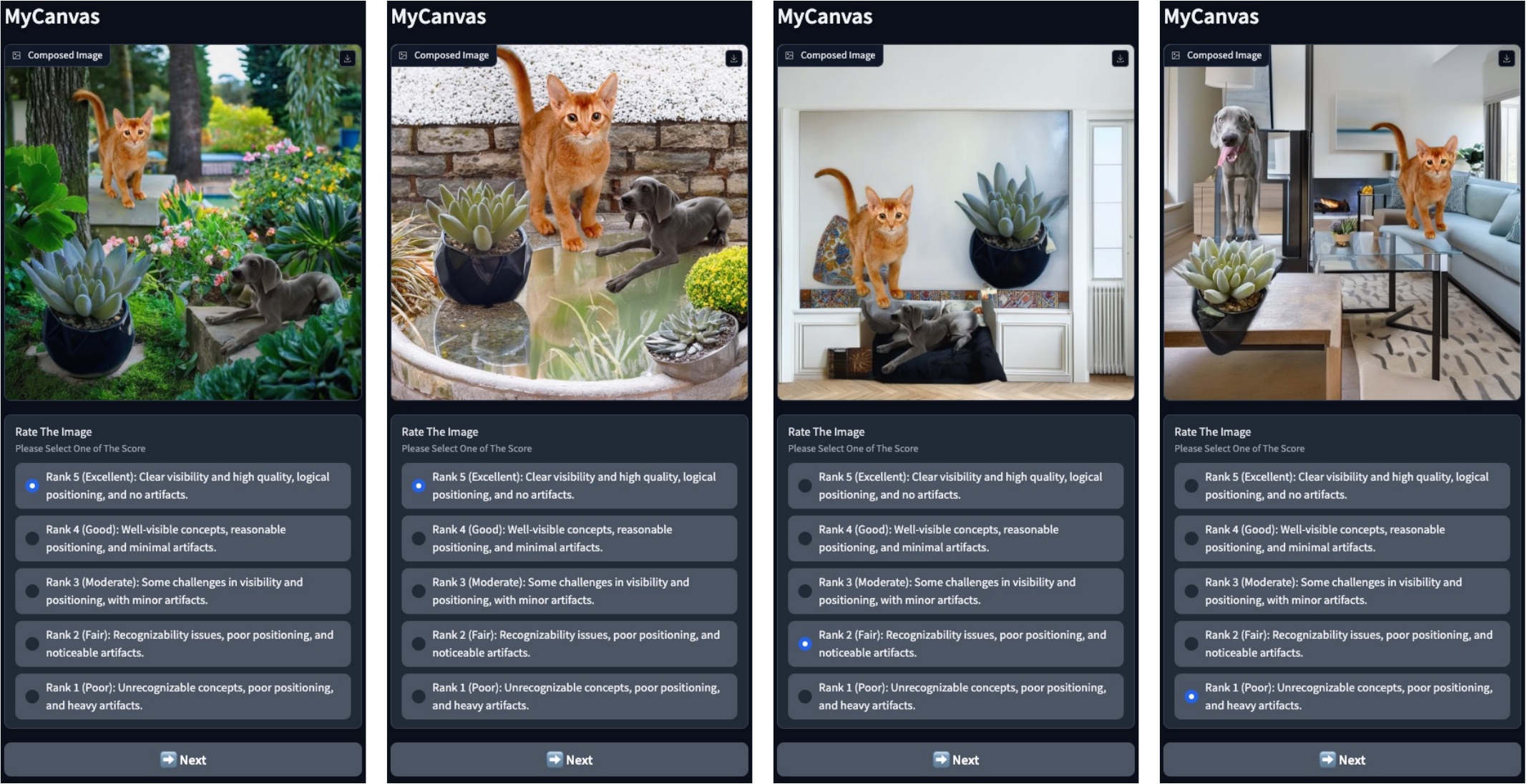}
    \caption{\textbf{A Evaluation Interface for \datasetname Data Filtering. } We demonstrate the evaluation interface for evaluating image quality and alignment.
    Users can rank the images from 1 to 5 by considering the following factors: 1) the presence of personalized concepts, 2) their reasonable positioning, and 3) the absence of artifacts.
    } 
    \label{fig:sup_interface}
\end{figure*}

\noindent\textbf{Evaluation Interface for Image Quality and Alignment.} As illustrated in Figure \ref{fig:sup_interface}, we have crafted an interface tailored for assessing both image quality and image-text alignment. Users have the straightforward option of expressing a preference by ranking the image from 1 to 5. Our evaluation criteria for image quality encompass three key considerations: 1) ensuring the presence of all specified concepts in the image, 2) validating the reasonable positioning of concepts, and 3) confirming the absence of artifacts on the image or the concept objects.

\noindent\textbf{Template for LLM-Guided Background Prompt.}

\begin{enumerate}
    \item We first leverage LLM (ChatGPT~\cite{openai2023gpt4}) 
 to generate background prompt by following the template provided by~\cite{lian2023llmgrounded}.
    \begin{bkg_template}
            You are an intelligent scene generator.
    I will provide you with a caption for a photo, image, or painting.
    Your task is to generate the background scene for the objects mentioned in the caption.
    \end{bkg_template}
    
    \item We then provide few curated examples to clarify the output format after the task description.
    \begin{bkg_query}

    \textbf{Caption:} A photo of a green car parking on the left of a blue truck, with a red air balloon and a bird in the sky \\
    \textbf{Scene:} garden, forest, grass \\
    \textbf{Background:} in the garden, in the forest, on the grass \\

    \textbf{Caption:} A painting of a wooden table in the living room with an apple on it \\
    \textbf{Scene:} room, kitchen, office \\
    \textbf{Background:} in the room, in the kitchen, in the office \\
    $\cdots$ \\
    \textbf{Caption:} A black refrigerator in a newly decorated house. \\
    \textbf{Scene:} kitchen, room, office \\
    \textbf{Background:} in the kitchen, in the room, in the office \\
    \end{bkg_query}
    
    \item We prompt the composition (e.g., car, cat, dog, house) and obtain three possible backgrounds from LLM.
    \begin{bkg_prompt}
        \textbf{Text prompt:} one car, one cat, one dog and one house \\
        \textbf{The background prompt:} on the street,in the suburban neighborhood,in the countryside
    \end{bkg_prompt}
\end{enumerate}

\noindent\textbf{Template for LLM-Guided Object Composition Layout.}

\begin{enumerate}
    \item Define the target task and the desired layout format.

    \begin{fs_template}
        You are an intelligent bounding box generator. \\\\
        I will provide you with a caption for a photo, image, or painting. \\\\
        Your task is to generate the bounding boxes for the objects mentioned in the caption, along with a background prompt describing the scene. \\\\
        The images are of size 512x512, and the bounding boxes should not overlap or go beyond the image boundaries. \\\\
        Each bounding box should be in the format of (object name, [top-left x coordinate, top-left y coordinate, box width, box height]) and include exactly one object. \\
        Make the boxes larger if possible. \\\\
        Do not put objects that are already provided in the bounding boxes into the background prompt. \\\\
        If needed, you can make reasonable guesses. \\\\
        Generate the object descriptions and background prompts in English even if the caption might not be in English. \\\\
        Do not include non-existing or excluded objects in the background prompt. \\\\
        Please refer to the example below for the desired format. \\\\
        Please note that a dialogue box is also an object. \\
        \\
        MAKE A REASONABLE GUESS OBJECTS MAY BE IN WHAT PLACE. \\
        The top-left x coordinate + box width MUST NOT BE HIGHER THAN 512. \\
        The top-left y coordinate + box height MUST NOT BE HIGHER THAN 512. \\
        ENSURE that generated bounding boxes NOT overlapped with each other.
    \end{fs_template}
    \item We then provide relevant examples that assist the LLM in understanding context and format requirements.
    \begin{fs_query}
        \textbf{Caption:} A couple of people with a laptop at a table. \\
        \textbf{Objects:} [["keyboard", [141, 309, 197, 117]]] \\

        \textbf{Caption:} Full course dinner served on large plate including drinks and dessert. \\
        \textbf{Objects:} [["cup", [0, 69, 92, 165]], ["cup", [18, 10, 96, 137]], ["fork", [165, 326, 61, 168]], ["carrot", [95, 308, 84, 71]], ["cake", [244, 41, 91, 93]]] \\

        \textbf{Caption:} A white plate served on a multi colored table cloth \\
        \textbf{Objects:} [["fork", [462, 120, 36, 248]], ["knife", [0, 162, 74, 212]], ["orange", [287, 129, 92, 100]], ["broccoli", [199, 163, 186, 170]]] \\
        
        $\cdots$ \\
    
        \textbf{Caption:} Two plastic containers next to a banana on a table. \\
        \textbf{Objects:} [["apple", [391, 207, 134, 69]], ["apple", [376, 264, 167, 187]], ["pizza", [124, 79, 202, 118]], ["pizza", [81, 128, 213, 120]], ["bowl", [366, 189, 202, 270]], ["sandwich", [86, 124, 215, 128]]]

    \end{fs_query}
    \item Finally, input the text prompt for layout generation. LLM generates the layout in response to the given query.
    \begin{prompt}
        \textbf{Text prompt:} one car, one cat, one dog and one house on the street. \\
        \textbf{Returned Objects:} ([('car', [0, 960, 836, 1408]), ('cat', [1364, 1476, 1856, 1864]), ('dog', [280, 1460, 880, 2048]), ('house', [960, 772, 2048, 2016])])
    \end{prompt}
\end{enumerate}

\section{Experiments}
\label{sec:supp-experiment}
In this section, we show more experiments, including ablation and analysis of our method.

\begin{figure*}[!t]
    \centering
    \includegraphics[width=\linewidth]{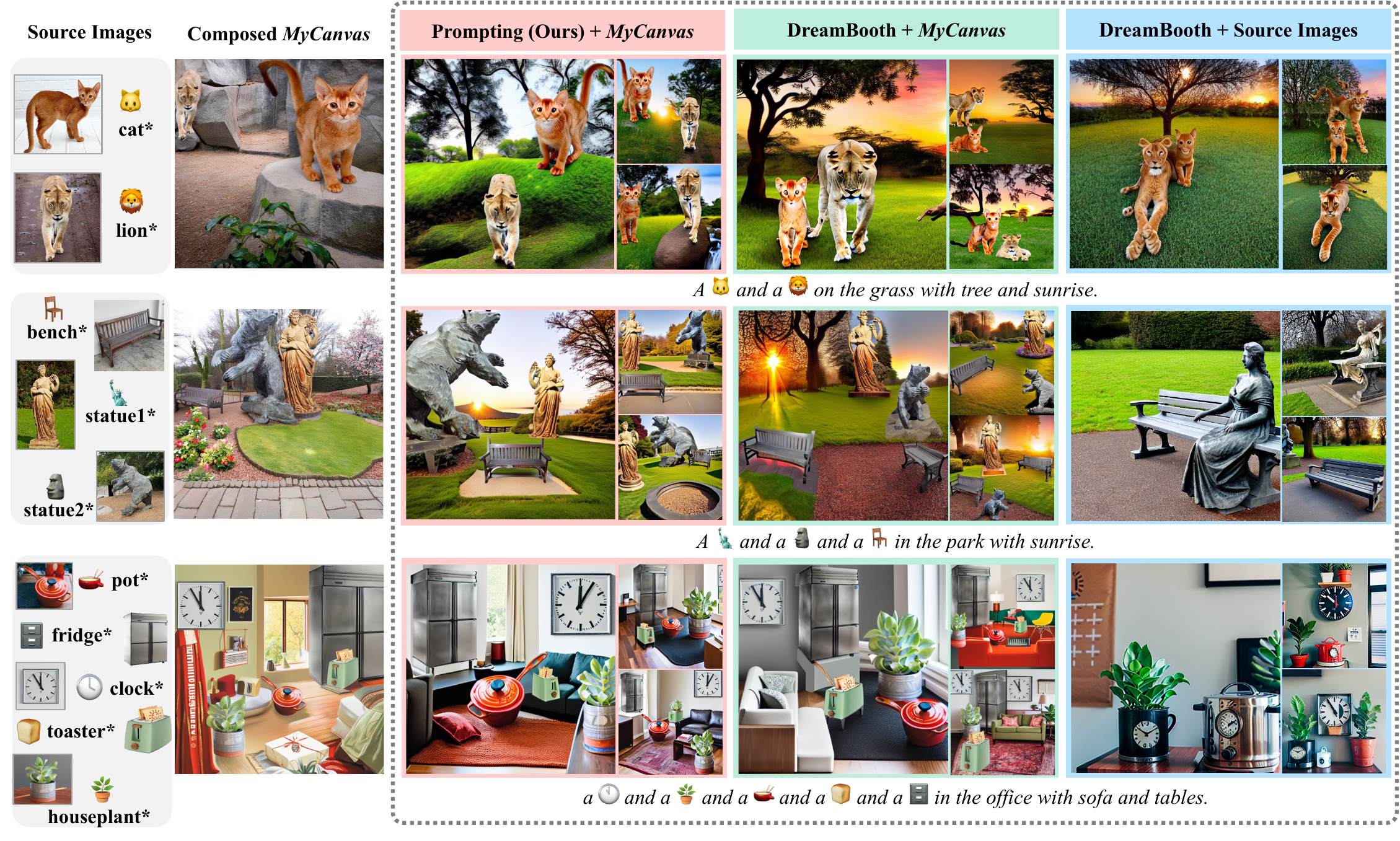}

    \caption{\textbf{Qualitative Results for DreamBooth Integrated with Our \datasetname. } This figure showcases a qualitative evaluation of the enhancement achieved by integrating our \datasetname into DreamBooth. 
    Much like the observed improvements with Custom Diffusion, our \datasetname exhibits significant advancements. 
    } 
    \label{fig:sup_gen_dreambooth}

\end{figure*}

\noindent\textbf{\datasetname Dataset on DreamBooth~\cite{ruiz2023dreambooth}.} In Figure~\ref{fig:sup_gen_dreambooth}, we present a qualitative comparison with DreamBooth. Similar to the observed phenomenon with Custom Diffusion~\cite{kumari2023multi}, our \datasetname dataset shows significant improvements when combined with DreamBooth. It excels in disentangling object identities that share similarities in the latent space, such as distinguishing between a cat and a lion. The incorporation of our prompting strategy yields additional improvements in aligning captions during image generation, ensuring an accurate representation of all objects. This is particularly evident in more challenging scenarios, such as compositions involving five objects. The results demonstrate the adaptability of our \datasetname to various personalized text-to-image diffusion models and its capacity to bring about improvements in generation quality.

\section{Personalized Composition Metrics}
\label{sec:supp-metrics}
In this section, we show more details about the metrics. In particular, we provide some brief intuitions regarding how to interpret our CP-CLIP and TI-CLIP.

\noindent\textbf{Score Interpretability.}\label{score_interpret} It is important to note that while both of our derived scores have the theoretical bounds of $ 0 \leq \mbox{CP-CLIP, TI-CLIP} \leq 1 $, it is practically impossible to reach the maximum. For CP-CLIP, we compute the average CLIP score. Even when what we generate is exactly the same as one of the instances in the source image, it would be different from other images, thus leading to an average smaller than 1. We empirically see that two images of the same object from different angles generate a CP-CLIP score of around 0.6 to 0.7.

On the other hand, enforcing a maximum of 1 over TI-CLIP is forcing completely identical features from two modalities. Also, TI-CLIP takes the entire image but only compares it against the background text prompt; when the foreground is occupying more space, this score may also likely decrease. Therefore, a better personalization model should see an increase in CP-CLIP but a \textbf{maintenance in TI-CLIP} to show that it is not overfitting.

\section{Implementation Details}
\label{sec:supp-implement-details}
Our training setup adheres to the configuration established by Custom Diffusion~\cite{kumari2023multi} as a foundational base. However, we make few modifications when conducting multi-concept experiments. All experiments are executed using Stable Diffusion 1.5~\cite{rombach2022high}, with a batch size of 8 (batch size of 2 for a single GPU, and typically leveraging 4 GPUs) and a learning rate of $10^{-5}$. Notably, for compositions involving 2 to 3 concepts, we observe that training for 2000 steps yields satisfactory image generation quality concerning compositionality and concept fidelity. In instances where compositions include 4 to 5 concepts, we extend the training to 3000 steps in our experiments. As for the regularization set, we employ 70 images for each concept category.

\end{document}